\documentclass{article}

\PassOptionsToPackage{square,numbers,comma,sort}{natbib}

\usepackage[final]{neurips_2025}

\usepackage{lipsum}
\usepackage{array}
\usepackage{times}
\usepackage{epsfig}
\usepackage{graphicx}
\usepackage{float}
\usepackage{wrapfig}
\usepackage{amsmath,amssymb,amsthm}
\usepackage{algorithm,algorithmicx,algpseudocode}
\usepackage{bm,xspace}
\usepackage{comment}
\usepackage{multirow}
\usepackage{balance}
\usepackage{url}
\usepackage{booktabs}
\usepackage{etoolbox,siunitx}
\usepackage{calc}
\usepackage{pifont,hologo}
\usepackage{color}
\usepackage{adjustbox}
\usepackage{amsmath}
\usepackage{enumitem}
\usepackage{subcaption}
\usepackage{bbding}  %
\usepackage[normalem]{ulem}  %
\usepackage{contour}
\usepackage[table]{xcolor}
\usepackage{soul}%
\usepackage{tocloft} %
\usepackage{etoc} %
\PassOptionsToPackage{dvipsnames}{xcolor}
\usepackage{algorithm}
\usepackage{algpseudocode}
\usepackage{listings}
\usepackage{caption} %
\usepackage[pagebackref,breaklinks,colorlinks,linkcolor=link,citecolor=cite,breaklinks=true]{hyperref}

\newcommand{\authorskip}{\hspace{2.5mm}}
\newcommand{\institutionskip}{\hspace{5.0mm}}

\definecolor{darkblue}{HTML}{35394B}
\definecolor{blue}{HTML}{004bb3}
\definecolor{red}{HTML}{cc1100}
\definecolor{orange}{HTML}{cc7700}
\definecolor{gray}{HTML}{efefef}
\definecolor{darkgreen}{HTML}{228B22}
\definecolor{darkgray}{HTML}{808080}
\definecolor{lightpurple}{HTML}{a56dba}

\definecolor{cite}{HTML}{3270b5}
\definecolor{link}{HTML}{b53532}
\definecolor{link}{HTML}{cc1100}
\definecolor{scratch}{HTML}{001219}
\definecolor{pretrain}{HTML}{0a9396}

\newcommand{\scratch}{\textcolor{scratch}{$\mathbf{\circ}$\,}}
\newcommand{\pretrain}{\textcolor{pretrain}{$\bullet$\,}}

\contourlength{0.8pt}

\newcommand{\figref}[1]{Fig.~\ref{#1}}
\newcommand{\tabref}[1]{Tab.~\ref{#1}}
\newcommand{\secref}[1]{Sec.~\ref{#1}}
\renewcommand{\eqref}[1]{Eq.~\ref{#1}}

\newcolumntype{x}[1]{>{\centering\arraybackslash}p{#1}}
\newcolumntype{y}[1]{>{\raggedright\arraybackslash}p{#1}}
\newcolumntype{z}[1]{>{\raggedleft\arraybackslash}p{#1}}
\newcommand{\tablestyle}[2]{\setlength{\tabcolsep}{#1}\renewcommand{\arraystretch}{#2}\centering\fontsize{8pt}{9pt}\selectfont}

\DeclareMathSymbol{@}{\mathord}{letters}{"3B}

\newcommand\mypara[1]{\vspace{0mm}\noindent\textbf{#1}}

\newcommand\tinyless{\raisebox{0.3ex}{\tiny $<$}}

\makeatletter
\DeclareRobustCommand\onedot{\futurelet\@let@token\@onedot}
\def\@onedot{\ifx\@let@token.\else.\null\fi\xspace}

\newcommand*{\Rom}[1]{\expandafter\@slowromancap\romannumeral #1@}
\newcommand*{\rom}[1]{\expandafter\romannumeral #1}

\def\1{\bm{1}}

\def\vp{{\bm{p}}}

\def\vt{{\bm{t}}}

\def\mD{{\bm{D}}}

\def\mK{{\bm{K}}}

\def\mR{{\bm{R}}}

\DeclareMathAlphabet{\mathsfit}{\encodingdefault}{\sfdefault}{m}{sl}
\SetMathAlphabet{\mathsfit}{bold}{\encodingdefault}{\sfdefault}{bx}{n}

\def\sP{{\mathcal{P}}}

\let\originalleft\left
\let\originalright\right
\renewcommand{\left}{\mathopen{}\mathclose\bgroup\originalleft}
\renewcommand{\right}{\aftergroup\egroup\originalright}

\usepackage{enumitem} %
\setlist{nosep, leftmargin=2em}

\usepackage[utf8]{inputenc} %
\usepackage[T1]{fontenc}    %
\usepackage{hyperref}       %
\usepackage{url}            %
\usepackage{booktabs}       %
\usepackage{amsfonts}       %
\usepackage{nicefrac}       %
\usepackage{microtype}      %
\usepackage{changepage,threeparttable} %

\title{Concerto: Joint 2D-3D Self-Supervised Learning Emerges Spatial Representations}

\author{Yujia Zhang\textsuperscript{\mdseries1} \authorskip 
Xiaoyang Wu\textsuperscript{\mdseries1$\dag$} \authorskip
Yixing Lao\textsuperscript{\mdseries1} \authorskip
Chengyao Wang\textsuperscript{\mdseries2} \\
\textbf{Zhuotao Tian}\textsuperscript{\mdseries3} \authorskip
\textbf{Naiyan Wang} \authorskip
\textbf{Hengshuang Zhao}\textsuperscript{\mdseries1$\ddag$} \authorskip
 \\
\textsuperscript{1}The University of Hong Kong \institutionskip
\textsuperscript{2}The Chinese University of Hong Kong \\
\textsuperscript{3}Harbin Institute of Technology (Shenzhen)\\ 
\vspace{1mm}
  \small{\textsuperscript{$\dag$}project lead\hspace{0.5cm}\textsuperscript{$\ddag$}corresponding author}\\
{\tt\small \url{https://pointcept.github.io/Concerto}}
}

\begin{document}
\vspace*{-15mm}
\maketitle
\vspace{-6mm}
\begin{center}
    \captionsetup{type=figure}
    \includegraphics[width=\linewidth]{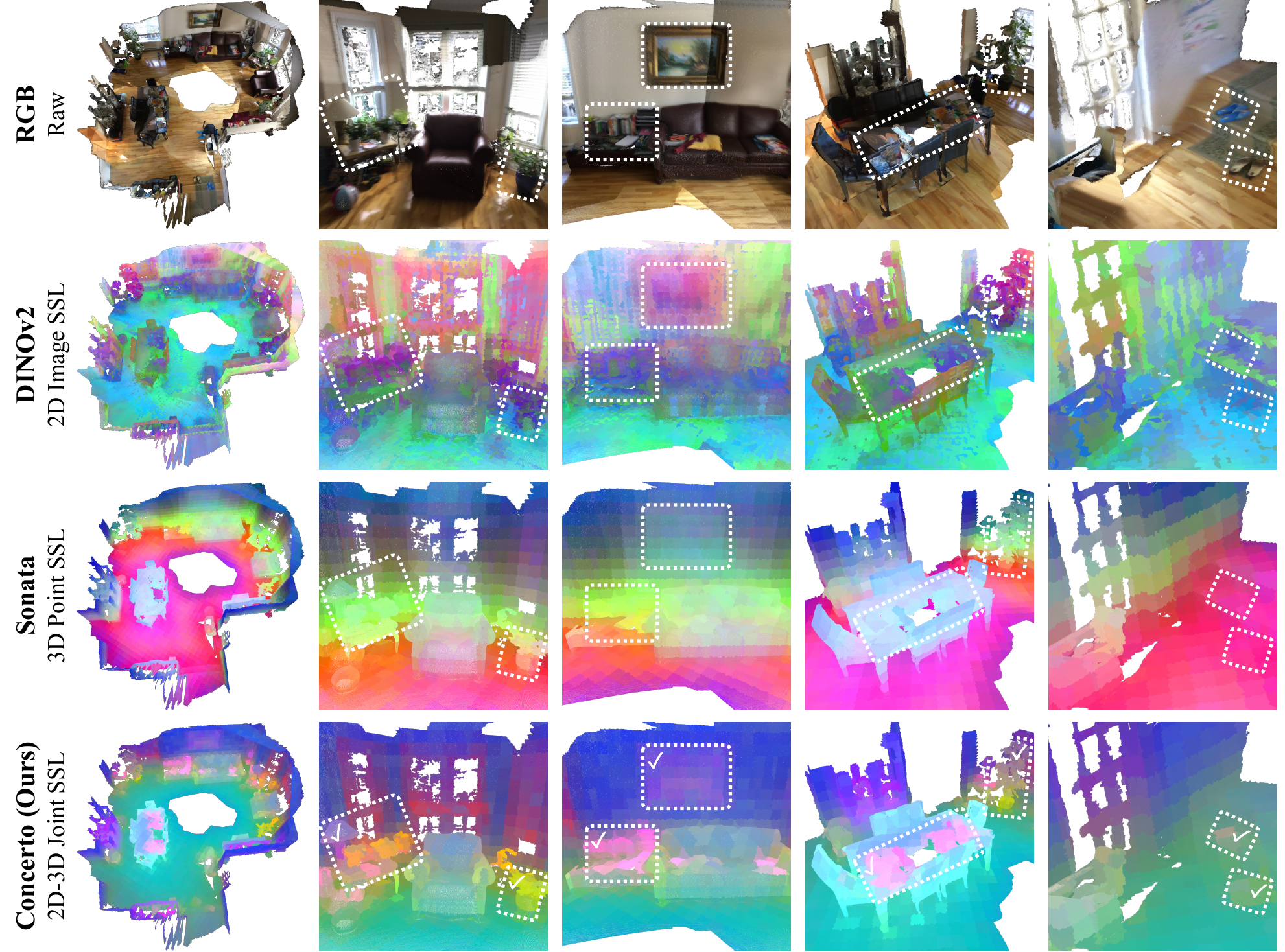}
    \vspace{-6mm}
    \caption{
   We visualize the principal components of point features learned by 2D and 3D self-supervised models~\cite{oquab2023dinov2,wu2025sonata}, mapped to RGB colors. DINOv2 lacks geometric awareness, and Sonata struggles to capture fine textures. Concerto integrates intra-modal self-distillation with cross-modal joint embedding prediction, enabling a self-supervised point cloud transformer~\cite{wu2024ptv3} to learn richer, emerging spatial representations with fine-grained geometric and semantic consistency across views.
    }\label{fig:teaser}
    \vspace{-1mm}
\end{center}

\begin{abstract}
    \vspace{-1mm}
    \vspace{-1mm}

Humans learn abstract concepts through multisensory synergy, and once formed, such representations can often be recalled from a single modality. Inspired by this principle, we introduce Concerto, a minimalist simulation of human concept learning for spatial cognition, combining 3D intra-modal self-distillation with 2D-3D cross-modal joint embedding. Despite its simplicity, Concerto learns more coherent and informative spatial features, as demonstrated by zero-shot visualizations. It outperforms both standalone SOTA 2D and 3D self-supervised models by 14.2\% and 4.8\%, respectively, as well as their feature concatenation, in linear probing for 3D scene perception. With full fine-tuning, Concerto sets new SOTA results across multiple scene understanding benchmarks (e.g., 80.7\% mIoU on ScanNet). We further present a variant of Concerto tailored for video-lifted point cloud spatial understanding, and a translator that linearly projects Concerto representations into CLIP’s language space, enabling open-world perception.
These results highlight that Concerto emerges spatial representations with superior fine-grained geometric and semantic consistency.

\end{abstract}

\vspace{-5mm}
\section{Introduction}
\label{sec:intro}
\vspace{-2mm}

Learning strong spatial representations in a self-supervised manner is foundational for spatial cognition tasks, spanning from low-level machine perception to high-level reasoning in domains such as autonomous driving~\cite{sun2020waymo,caesar2020nuscenes}, mixed reality~\cite{dehghan2021arkitscenes,james2019rlbench,straub2024efm3d}, and robotics~\cite{gu2023maniskill2}. Recent advances in self-supervised learning have significantly improved foundational representation models for the two dominant spatial modalities: 2D images~\cite{chen2020simclr,he2020moco,zhang2022dino,assran2023jpea, xie2022simmim} and 3D point clouds~\cite{xie2020pointcontrast,Pang2022pointmae,yu2021pointbert,wu2023msc,wu2025sonata}. Without the need for human annotations, these models have demonstrated strong performance across various downstream tasks by enabling the learning of geometry, and semantics at scale.

However, despite their individual successes within each data modality, our pilot study reveals that self-supervised representations learned independently from images and point clouds do not fully overlap. Specifically, concatenating features from self-supervised image models (e.g., DINOv2~\cite{oquab2023dinov2}) and point cloud models (e.g., Sonata~\cite{wu2025sonata}) leads to improved linear probing performance, suggesting that each modality captures complementary, rather than redundant, aspects of spatial information. The observation hints at the existence of \textit{a more robust and rich feature space that emerges from the interaction between 2D and 3D modalities}, indicating the core aim of this research: \textit{to uncover superior spatial representations through multi-modal self-supervised learning}.

\begin{wrapfigure}{r}{0.25\linewidth}
\vspace{-6mm}
  \begin{center}
    \includegraphics[width=\linewidth]{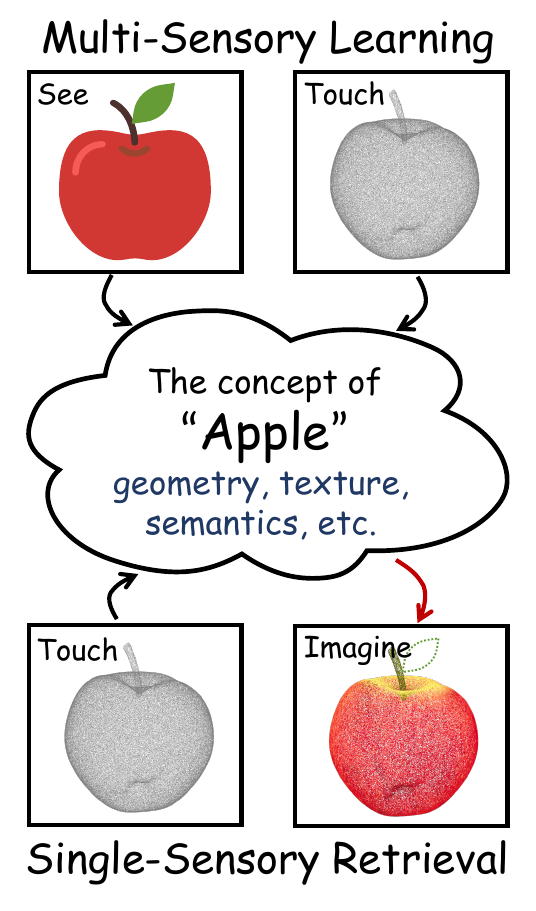}
  \end{center}
  \vspace{-3mm}
  \caption{\textbf{The ``Apple'' concept in cognition.}}
  \label{fig:apple}
  \vspace{-5mm}
\end{wrapfigure}

Our inspiration toward this target is rooted in how humans learn abstract concepts: through multisensory synergy~\cite{barsalou2008grounded, shams2008benefits}. Consider the example of an \textit{apple} (as illustrated in \figref{fig:apple})—our understanding of it is not limited to its visual appearance. Instead, the concept is formed through repeatedly seeing, touching, and tasting apples, allowing us to internalize its geometry, texture, and semantic meaning in a unified, predictive way. This cognitive process reflects a continuous, cross-modal integration of sensory data into a grounded concept space (\figref{fig:apple} top). Yet once such a representation is formed, it can be evoked from just a single modality: seeing an image of an apple can vividly recall its weight and texture, just as holding one can bring its color and shape to mind (\figref{fig:apple} bottom). This ability to retrieve rich, structured knowledge from partial sensory input underscores the importance of learning modality-agnostic representations that are both unified and predictive.

Driven by this vision, our methodology aims to offer a simple yet effective imitation of human multisensory synergy. To this end, we compose a \textbf{Concerto} of 2D-3D joint self-supervised learning, coupling intra-modal self-distillation for point representations~\cite{oquab2023dinov2,wu2025sonata} with cross-modal joint embedding prediction from images to point clouds~\cite{lecun2022path}. Ultimately, this training yields a self-supervised PTv3~\cite{wu2024ptv3} model, pretrained on 40k raw point clouds and 300k images~\cite{dai2017scannet,yeshwanth2023scannet++,armeni2016s3dis,dehghan2021arkitscenes,ramakrishnan2021hm3d,zheng2020structured3d,avetisyan2024scenescript}. In addition, we present a \textit{variation} of Concerto, augmented with an additional set of 50k point clouds with 200k corresponding images lifted from scene videos~\cite{zhou2018re10k} via feed-forward reconstruction~\cite{wang2025vggt}, tailored for video-based spatial understanding. We also introduce an \textit{interlude} of Concerto: a learned translator that linearly projects self-supervised representations into CLIP's language space~\cite{radford2021clip,li2022lseg}, enabling open-world perception.

The intersection of 2D and 3D self-supervised learning in Concerto yields a powerful synergy, enabling the emergence of superior spatial representations. PCA-colored visualizations reveal that Concerto captures more coherent and informative spatial features than SOTA 2D or 3D self-supervised models trained on a single modality (see \figref{fig:teaser}). Concerto exceeds its predecessor, Sonata, with a 4.8\% improvement in linear probing, achieving 77.3\% mIoU on ScanNet semantic segmentation using a single linear layer. Notably, this performance also surpasses the concatenation of Sonata and DINOv2 (1.4\%), demonstrating that the multisensory synergy in Concerto exceeds the representational upper bound achievable by single-modality self-supervised learning. With full fine-tuning, Concerto achieves SOTA performance across a range of scene perception tasks. For example, reaching 80.7\% mIoU on ScanNet semantic segmentation.

\newpage

\begin{table}
\centering
    \tablestyle{7.6pt}{1.08}
        \begin{tabular}{lll|rrrrrrr}\toprule
\multicolumn{3}{c|}{Semantic Segmentation} &\multicolumn{3}{c}{ScanNet Val~\cite{dai2017scannet}} &\multicolumn{3}{c}{ScanNet200 Val~\cite{rozenberszki2022scannet200}} \\
\cmidrule(lr){1-3}\cmidrule(lr){4-6}\cmidrule(lr){7-9}
Method &Type &Encoder &mIoU &mAcc &allAcc &mIoU &mAcc &allAcc \\\midrule
DINOv2~\cite{oquab2023dinov2} &2D Image SSL  &ViT-G &63.09 &75.50 &82.42 &27.42 &37.59 &72.80 \\
Sonata~\cite{wu2025sonata} &3D Point SSL &PTv3-B &72.52 &83.11 &89.74 &29.25 &41.61 &81.15 \\
Sonata$\times$DINOv2 &3D SSL$\times$2D SSL &Both &75.91 &85.36 &91.25 &36.67 &46.98 &82.85 \\
    \cellcolor[HTML]{f3f7fc}\textbf{Concerto}~(ours) &\cellcolor[HTML]{f3f7fc}2D-3D Joint SSL &\cellcolor[HTML]{f3f7fc}PTv3-B &\textbf{77.32} &\textbf{86.58} &\textbf{91.74} &\textbf{37.41} &\textbf{49.49} &\textbf{83.29} \\
\bottomrule
\end{tabular}

        \vspace{1mm}
        \caption{\textbf{Linear probing results on 3D semantic segmentation.} We compare self-supervised features learned from 2D, 3D, their feature concatenation, and our 2D-3D joint SSL model, Concerto (as a preview). Notably, the concatenation of 2D and 3D features outperforms either modality alone, suggesting that the two modalities encode complementary information.
        Concerto achieves the best performance across all metrics, demonstrating its ability to learn superior spatial representations.}
        \label{tab:2dx3d}
        \vspace{-6mm}
\end{table}

\vspace{-1mm}
\section{Beyond Single Modality: Toward a New World of Representations}
\vspace{-1mm}

This section presents a pilot study to explore high-level questions surrounding self-supervised representations. These questions form the conceptual foundation of our research, and our methodology emerges as a natural and simple response to the insights gained here.

\vspace{-1mm}
\subsection{Is There a Superior Representation Space Beyond Single-Modality Learning?}
\vspace{-1mm}

Self-supervised learning on 2D images and 3D point clouds has achieved remarkable progress in visual representation learning. However, when trained independently, these models may capture only modality-specific perspectives of the spatial world. Just as a person who has only seen an \textit{apple} but never tasted one may lack a sense of its flavor or texture, single-modal learning inevitably misses critical dimensions of the world. This raises a fundamental question: Is there a superior representation space that can emerge from the synergy between 2D and 3D modalities?

To probe this question, we begin with a simple pilot experiment: fusing self-supervised features from image and point cloud models, prior to any explicit learning of cross-modal synergy. Specifically, we select two representative self-supervised models trained independently on images (DINOv2~\cite{oquab2023dinov2}) and point clouds (Sonata~\cite{wu2025sonata}). We lift image features into 3D space using depth and camera parameters, and concatenate them with point cloud representations to enable feature-level fusion. We benchmark the 2D, 3D, and fused representations via linear probing on 3D scene-level semantic segmentation using the ScanNet~\cite{dai2017scannet} dataset, with results presented in \tabref{tab:2dx3d}. Notably, this naive combination outperforms both individual modalities, suggesting the presence of complementary information and hinting at a richer representational space.

However, simply concatenating 2D and 3D self-supervised features, while yielding a stronger representation space, still falls short of uncovering the \textit{unexplored new world} we are seeking. This approach lacks integration during learning and cannot fully capture the synergy that emerges when modalities are learned together. The deeper insight lies in the potential of multi-modal joint representation learning---not only to align complementary signals across modalities, but also to form coherent, predictive embeddings that generalize beyond their source. Ideally, such fused representations can be retrieved from a single modality, even if they were originally learned through multi-modal interaction. This form of joint 2D-3D representation learning is intuitive, as it mirrors how humans form concepts, as discussed in \secref{sec:intro} and illustrated in \figref{fig:apple}.

\textit{This insight leads to our methodology: designing a unified framework that learns to embed spatial information through both intra-modal refinement and cross-modal prediction.}

\vspace{-1mm}
\subsection{Can Multi-modal Self-Supervised Representations Speak the Language of Concepts?}
\vspace{-1mm}

Human language is often considered a compressed and symbolic interface to abstract concepts learned through multisensory synergy~\cite{barsalou2008grounded}. If multi-modal self-supervised representation learning succeeds in forming unified abstract concepts, then such representations should, in principle, be able to align with human language---perhaps even through a simple linear projection. This perspective raises a natural question: Can multi-modal self-supervised representations, learned entirely without human language, speak the language of concepts?

We believe the answer to this question is ultimately yes. However, our current study is grounded in the spatial domain, leveraging only 2D images and 3D point clouds. This limited sensory scope makes it challenging to fully align with human language, which emerges from a far richer blend of modalities. Still, we propose that linearly probing self-supervised representations into a language embedding space, such as CLIP's, offers a meaningful way to evaluate progress toward this goal. Beyond serving as a diagnostic tool, this projection also extends the open-vocabulary capabilities of self-supervised spatial representations, offering a step toward broader concept grounding.

\textit{We propose linear probing into the CLIP feature space as a next-level evaluation criterion for self-supervised learning beyond single modality.}

\vspace{-1mm}
\section{Concerto: Joint 2D-3D Self-Supervised Learning}
\vspace{-1mm}

\begin{figure}
  \centering
  \includegraphics[width=1.0\textwidth]{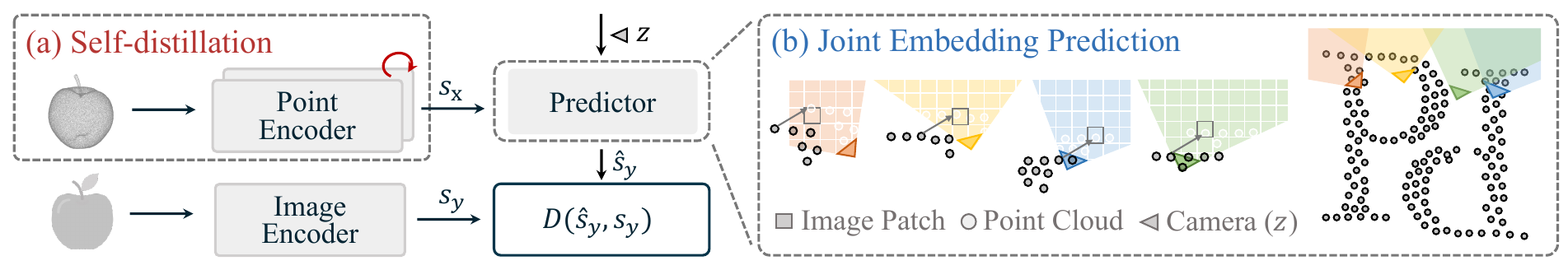}
  \caption{\textbf{Overview of the Concerto architecture.} Concerto simulates human multisensory synergy by coupling (a) intra-modal self-distillation on 3D point clouds to progressively refine its internal spatial representations (see \secref{sec:intra}), and (b) cross-modal joint embedding prediction that aligns point features with corresponding image patch features using camera parameters (see \secref{sec:cross}). The self-distillation branch (a) employs a restricted online clustering objective, while the joint embedding prediction (b) applies a looser cosine similarity constraint. This dual self-supervised objective encourages the emergence of coherent, modality-agnostic spatial representations.}
  \vspace{-2mm}
  \label{fig:method}
\end{figure}

This section introduces the joint 2D-3D self-supervised learning framework of Concerto. The proposed architecture is intentionally simple, designed to highlight the power of multisensory synergy through strong empirical performance. An overview of concerto architecture is present in \figref{fig:method}.

\vspace{-1mm}
\subsection{Intra-Modal Self-Distillation}
\label{sec:intra}
\vspace{-1mm}

The primary duty of Concerto is to validate the superiority of multi-modal synergy, rather than seeking innovation in single-modality self-supervised learning architecture. For this reason, the intra-modal branch of Concerto builds upon the recent Sonata framework~\cite{wu2025sonata}, which applies self-distillation~\cite{caron2021emerging, oquab2023dinov2} to learn point cloud representations without supervision. We briefly revisit this architecture here and refer readers to the original Sonata paper for a detailed discussion on building a reliable self-supervised framework for sparse and unstructured point cloud data.

The intra-modal self-distillation in Concerto focuses solely on the 3D domain, where a Point Transformer V3~\cite{wu2024ptv3} is trained to produce stable and predictive features through a teacher-student paradigm. The student encoder is optimized to match the output of a momentum-updated teacher~\cite{he2020moco}, using a clustering-based objective~\cite{caron2020swav} that promotes consistency across augmented views of the same point cloud. A unique challenge in sparse point clouds is the geometric shortcut, where models collapse to easily accessible low-level geometric cues. These cues are not learned but introduced implicitly through the local kernel definitions of point cloud operators. Sonata mitigates this through several micro-designs that obscure explicit spatial signals and encourage learning from input features. This self-motivated refinement process allows the model to internalize geometric and structural priors from 3D data, forming the foundation for multi-modal learning in Concerto.

\vspace{-1mm}
\subsection{Cross-Modal Joint Embedding Prediction}
\label{sec:cross}
\vspace{-1mm}

We introduce an additional cross-modal self-supervised objective that continuously stimulates synergy from the image’s self-supervised representation into the point cloud domain. This design aligns with the core vision of LeCun’s Joint Embedding Predictive Architecture (JEPA)~\cite{lecun2022path}, which advocates learning by predicting latent representations across modalities using a conditional predictor. The goal is to predict point cloud embeddings that match the associated pixel embeddings extracted from a self-supervised image encoder (e.g., DINOv2~\cite{oquab2023dinov2}). Empirically, we find that cosine similarity provides the most effective criterion for training this predictive branch, and by applying strong point cloud data augmentations and exploring less aggressive image augmentations compared to DINOv2, Concerto learns more generalizable representations.

In our implementation, we divided scenes with a large number of images into several data pieces. Each one consists of one point cloud with 4 images. For scenes with fewer images (e.g., 5), we retain the original dataset divisions. As shown in \figref{fig:method}, we respectively obtain point representations $s_x$ and image representations $s_y$ from the point encoder and image encoder. In the context of 2D images and 3D point clouds, our predictor takes camera parameters as the condition $z$ to establish correspondences between image pixels and point cloud points. The details of this process are provided in the Appendix. Then, for each image patch, we compute the mean of the features of points falling within it to get predicted patch features $\hat{s}_y$ from point clouds. Finally, we compute the loss $D(s_y, \hat{s}_y)$ using cosine similarity. This process introduces 2D self-supervised features into 3D self-supervised learning and stimulates the point intra-modal self-distillation process.

\vspace{-1mm}
\subsection{Synergy Emerged from Joint Self-Supervised Learning}
\vspace{-1mm}
The combination of intra-modal self-distillation and cross-modal joint embedding in Concerto emerges as a strong synergy, surpassing single-modality learning and elevating the interaction between self-supervised point cloud and image representations beyond naive fusion. The complementary cognitive signals from image self-supervised learning, through cross-modal joint embedding prediction, encourage point intra-modal self-distillation to extend capacity beyond a single modality and emerge a superior spatial representation surpassing the naive combination of self-supervised features from images and point clouds. This process yields spatial representations that are more expressive than those obtained by merely concatenating features from separate 2D and 3D models. We believe this chain reaction is central to the design of Concerto. Notably, the architecture also supports training on point clouds without paired images, enabling hybrid self-supervised learning without compromising scalability on large-scale 3D datasets.

\vspace{-1mm}
\section{Experiments}
\vspace{-1mm}

We evaluate Concerto's representations with various scene perception tasks using the same protocols as Sonata~\cite{wu2025sonata}. Specifically, linear probing, which keeps the encoder frozen and adapts the upcasted original scale features to downstream tasks; decoder probing, adding a lightweight decoder after the frozen encoder to facilitate adaptation; and full fine-tuning, where the encoder and decoder are optimized for downstream tasks. We further analyze Concerto's key properties with these results.

\vspace{-1mm}
\subsection{Main Results}
\vspace{-2mm}

\begin{table*}[!t]
    \begin{minipage}{\textwidth}
    \centering
        \tablestyle{1.7pt}{1.08}
        \begin{tabular}{y{22.5mm}rz{8mm}cccccccccccc}\toprule
Fine-Tuning &\multicolumn{2}{c}{Params} &\multicolumn{3}{c}{ScanNet Val~\cite{dai2017scannet}} &\multicolumn{3}{c}{ScanNet200 Val~\cite{rozenberszki2022scannet200}} &\multicolumn{3}{c}{ScanNet++ Val~\cite{yeshwanth2023scannet++}} &\multicolumn{3}{c}{S3DIS Area 5~\cite{armeni2016s3dis}}\\
\cmidrule(lr){1-1} \cmidrule(lr){2-3} \cmidrule(lr){4-6} \cmidrule(lr){7-9} \cmidrule(lr){10-12} \cmidrule(lr){13-15}
Methods &\multicolumn{1}{c}{Learn.} &\multicolumn{1}{c}{Pct.} &mIoU &mAcc &allAcc &mIoU &mAcc &allAcc &mIoU &mAcc &allAcc &mIoU &mAcc &allAcc\\\midrule
\scratch \textcolor{darkgray}{SparseUNet}~\cite{choy20194d} &\textcolor{darkgray}{39.2M} &\textcolor{darkgray}{100\%} &\textcolor{darkgray}{72.3} &\textcolor{darkgray}{80.2} &\textcolor{darkgray}{90.0} &\textcolor{darkgray}{25.0} &\textcolor{darkgray}{32.9} &\textcolor{darkgray}{80.4} &\textcolor{darkgray}{28.8} &\textcolor{darkgray}{38.4} &\textcolor{darkgray}{80.1} &\textcolor{darkgray}{66.3} &\textcolor{darkgray}{72.5} &\textcolor{darkgray}{89.8} \\
\ \pretrain PC~\cite{xie2020pointcontrast} &39.2M &100\% &72.3 &80.9 &90.1 &26.2 &33.0 &79.9 &29.2 &39.7 &82.7 &68.1 &73.5 &90.0 \\
\ \pretrain CSC~\cite{hou2021csc} &39.2M &100\% &72.8 &81.0 &90.7 &26.9 &33.7 &80.6 &32.5 &41.1 &83.7 &70.7 &76.4 &90.8 \\
\ \pretrain MSC~\cite{wu2023msc} &39.2M &100\% &75.7 &83.4 &91.3 &32.0 &41.6 &82.3 &39.4 &49.6 &84.9 &70.7 &76.1 &91.0 \\
\cmidrule(lr){1-15}
\scratch \textcolor{darkgray}{PTv3}~\cite{wu2024ptv3} &\textcolor{darkgray}{124.8M} &\textcolor{darkgray}{100\%} &\textcolor{darkgray}{77.6} &\textcolor{darkgray}{85.0} &\textcolor{darkgray}{92.0} &\textcolor{darkgray}{35.3} &\textcolor{darkgray}{46.0} &\textcolor{darkgray}{83.4} &\textcolor{darkgray}{48.2} &\textcolor{darkgray}{61.6} &\textcolor{darkgray}{87.0} &\textcolor{darkgray}{73.4} &\textcolor{darkgray}{78.9} &\textcolor{darkgray}{91.7} \\
\ \pretrain MSC~\cite{wu2023msc} &124.8M &100\% &78.2 &85.3 &92.2 &33.4 &43.7 &83.4 &48.7 &61.9 &87.2 &69.9 &74.9 &91.2 \\
\ \pretrain Sonata~\cite{wu2025sonata} &124.8M &100\% &\cellcolor[HTML]{eef9f2}79.4 &\cellcolor[HTML]{eef9f2}86.1 &\cellcolor[HTML]{eef9f2}92.5 &\cellcolor[HTML]{eef9f2}36.8 &\cellcolor[HTML]{eef9f2}46.5 &\cellcolor[HTML]{eef9f2}84.4 &\cellcolor[HTML]{eef9f2}49.3 &\cellcolor[HTML]{eef9f2}62.4 &\cellcolor[HTML]{eef9f2}87.6 &\cellcolor[HTML]{eef9f2}76.0 &\cellcolor[HTML]{eef9f2}81.6 &\cellcolor[HTML]{eef9f2}93.0 \\
\cellcolor[HTML]{f3f7fc}\ \pretrain Concerto &124.8M &100\% &\cellcolor[HTML]{def3e6}\textbf{80.7} &\cellcolor[HTML]{def3e6}\textbf{87.4} &\cellcolor[HTML]{def3e6}\textbf{93.1} &\cellcolor[HTML]{def3e6}\textbf{39.2} &\cellcolor[HTML]{def3e6}\textbf{50.2} &\cellcolor[HTML]{def3e6}\textbf{85.0} &\cellcolor[HTML]{def3e6}\textbf{50.7} &\cellcolor[HTML]{def3e6}\textbf{63.3} &\cellcolor[HTML]{def3e6}\textbf{87.9} &\cellcolor[HTML]{def3e6}\textbf{77.4} &\cellcolor[HTML]{def3e6}\textbf{85.0} &\cellcolor[HTML]{def3e6}\textbf{93.2} \\
\bottomrule
\end{tabular}

        \vspace{-2mm}
        \subcaption{Full fine-tuning. We evaluate Concerto using full fine-tuning, unlocking both encoder and decoder, and compare semantic segmentation mIoU, mAcc, allAcc(\%) results across 4 benchmarks.}
        \label{subtab:semseg_fine_tuning}
        \vspace{1mm}
    \end{minipage} \\
    \begin{minipage}{\textwidth}
    \centering
        \tablestyle{1.7pt}{1.08}
        \begin{tabular}{y{22.5mm}rz{8mm}cccccccccccc}\toprule
Param. Efficiency &\multicolumn{2}{c}{Params} &\multicolumn{3}{c}{ScanNet Val~\cite{dai2017scannet}} &\multicolumn{3}{c}{ScanNet200 Val~\cite{rozenberszki2022scannet200}} &\multicolumn{3}{c}{ScanNet++ Val~\cite{yeshwanth2023scannet++}} &\multicolumn{3}{c}{S3DIS Area 5~\cite{armeni2016s3dis}}\\
\cmidrule(lr){1-1} \cmidrule(lr){2-3} \cmidrule(lr){4-6} \cmidrule(lr){7-9} \cmidrule(lr){10-12} \cmidrule(lr){13-15}
Methods &\multicolumn{1}{c}{Learn.} &\multicolumn{1}{c}{Pct.} &mIoU &mAcc &allAcc &mIoU &mAcc &allAcc &mIoU &mAcc &allAcc &mIoU &mAcc &allAcc\\\midrule
\scratch \textcolor{darkgray}{SparseUNet}~\cite{choy20194d} &\textcolor{darkgray}{39.2M} &\textcolor{darkgray}{100\%} &\textcolor{darkgray}{72.3} &\textcolor{darkgray}{80.2} &\textcolor{darkgray}{90.0} &\textcolor{darkgray}{25.0} &\textcolor{darkgray}{32.9} &\textcolor{darkgray}{80.4} &\textcolor{darkgray}{28.8} &\textcolor{darkgray}{38.4} &\textcolor{darkgray}{80.1} &\textcolor{darkgray}{66.3} &\textcolor{darkgray}{72.5} &\textcolor{darkgray}{89.8} \\
\ \pretrain PC~\cite{xie2020pointcontrast} (lin.) &\tinyless0.2M &\tinyless0.1\% &5.6 &9.7 &50.0 &0.5 &0.9 &40.3 &1.8 &3.1 &46.4 &11.4 &18.6 &52.3 \\
\ \pretrain CSC~\cite{hou2021csc} (lin.) &\tinyless0.2M &\tinyless0.1\% &12.6 &18.1 &64.2 &1.3 &2.1 &53.0 &2.8 &4.5 &53.6 &24.4 &32.0 &66.4 \\
\ \pretrain MSC~\cite{wu2023msc} (lin.) &\tinyless0.2M &\tinyless0.1\% &14.1 &20.3 &62.9 &1.5 &2.5 &53.6 &4.5 &6.6 &61.3 &27.9 &35.5 &71.1 \\
\cmidrule{1-15}
\scratch \textcolor{darkgray}{PTv3}~\cite{wu2024ptv3} &\textcolor{darkgray}{124.8M} &\textcolor{darkgray}{100\%} &\textcolor{darkgray}{77.6} &\textcolor{darkgray}{85.0} &\textcolor{darkgray}{92.0} &\textcolor{darkgray}{35.3} &\textcolor{darkgray}{46.0} &\textcolor{darkgray}{83.4} &\textcolor{darkgray}{48.2} &\textcolor{darkgray}{61.6} &\textcolor{darkgray}{87.0} &\textcolor{darkgray}{73.4} &\textcolor{darkgray}{78.9} &\textcolor{darkgray}{91.7} \\
\ \pretrain MSC~\cite{wu2023msc} (lin.) &\tinyless0.2M &\tinyless0.2\% &21.8 &32.2 &65.5 &3.3 &5.5 &57.5 &8.1 &11.9 &64.7 &32.1 &42.4 &70.9 \\
\ \pretrain Sonata~\cite{wu2025sonata} (lin.) &\tinyless0.2M &\tinyless0.2\% &\cellcolor[HTML]{eef9f2}72.5 &\cellcolor[HTML]{eef9f2}83.1 &\cellcolor[HTML]{eef9f2}89.7 &\cellcolor[HTML]{eef9f2}29.3 &\cellcolor[HTML]{eef9f2}41.6 &\cellcolor[HTML]{eef9f2}81.2 &\cellcolor[HTML]{eef9f2}38.9 &\cellcolor[HTML]{eef9f2}52.8 &\cellcolor[HTML]{eef9f2}84.3 &\cellcolor[HTML]{eef9f2}72.3 &\cellcolor[HTML]{eef9f2}81.2 &\cellcolor[HTML]{eef9f2}90.9 \\
\cellcolor[HTML]{f3f7fc}\ \pretrain Concerto (lin.) &\tinyless0.2M &\tinyless0.2\% &\cellcolor[HTML]{def3e6}\textbf{77.3} &\cellcolor[HTML]{def3e6}\textbf{86.6} &\cellcolor[HTML]{def3e6}\textbf{91.7} &\cellcolor[HTML]{def3e6}\textbf{37.4} &\cellcolor[HTML]{def3e6}\textbf{49.5} &\cellcolor[HTML]{def3e6}\textbf{83.3} &\cellcolor[HTML]{def3e6}\textbf{45.6} &\cellcolor[HTML]{def3e6}\textbf{60.5} &\cellcolor[HTML]{def3e6}\textbf{86.5} &\cellcolor[HTML]{def3e6}\textbf{73.5} &\cellcolor[HTML]{def3e6}\textbf{81.3} &\cellcolor[HTML]{def3e6}\textbf{90.9} \\
\cmidrule(lr){1-15}
\ \pretrain Sonata~\cite{wu2025sonata} (dec.) &16.3M &13\% &\cellcolor[HTML]{eef9f2}79.1 &\cellcolor[HTML]{eef9f2}86.6 &\cellcolor[HTML]{def3e4}\textbf{92.7} &\cellcolor[HTML]{eef9f2}33.5 &\cellcolor[HTML]{eef9f2}44.5 &\cellcolor[HTML]{eef9f2}84.1 &\cellcolor[HTML]{eef9f2}45.2 &\cellcolor[HTML]{eef9f2}57.4 &\cellcolor[HTML]{eef9f2}86.8 &\cellcolor[HTML]{eef9f2}74.5 &\cellcolor[HTML]{eef9f2}80.4 &\cellcolor[HTML]{def3e4}\textbf{92.6} \\
\cellcolor[HTML]{f3f7fc}\ \pretrain Concerto (dec.) &16.3M &13\% &\cellcolor[HTML]{def3e6}\textbf{79.5} &\cellcolor[HTML]{def3e6}\textbf{87.6} &\cellcolor[HTML]{eef9f0}92.6 &\cellcolor[HTML]{def3e6}\textbf{37.8} &\cellcolor[HTML]{def3e6}\textbf{50.5} &\cellcolor[HTML]{def3e6}\textbf{84.1} &\cellcolor[HTML]{def3e6}\textbf{48.3} &\cellcolor[HTML]{def3e6}\textbf{62.3} &\cellcolor[HTML]{def3e6}\textbf{87.7} &\cellcolor[HTML]{def3e6}\textbf{75.5} &\cellcolor[HTML]{def3e6}\textbf{84.2} &\cellcolor[HTML]{eef9f2}92.3 \\
\bottomrule
\end{tabular}

        \vspace{-2mm}
        \subcaption{Parameter efficiency. By using linear probing (lin.) and decoder probing (dec.), we compare semantic segmentation mIoU, mAcc, allAcc(\%) results across 4 benchmarks.}
        \label{subtab:semseg_param_efficiency}
        \vspace{1mm}
    \end{minipage} \\
    \begin{minipage}{\textwidth}
    \centering
        \tablestyle{8.8pt}{1.08}
        \begin{tabular}{y{22.5mm}rrrrrrrrrrr}\toprule
Data Efficiency &\multicolumn{5}{c}{Limited Scenes (Pct.)} &\multicolumn{5}{c}{Limited Annotation (Pts.)} \\
\cmidrule(lr){1-1} \cmidrule(lr){2-6} \cmidrule(lr){7-11}
Methods &1\% &5\% &10\% &20\% &Full &20 &50 &100 &200 &Full \\\midrule
\scratch PTv2~\cite{wu2022ptv2} &24.8 &48.1 &59.8 &66.3 &\textcolor{darkgray}{75.4} &58.4 &66.1 &70.3 &71.2 &\textcolor{darkgray}{75.4} \\
\scratch SparseUNet~\cite{choy20194d} &26.0 &47.8 &56.7 &62.9 &\textcolor{darkgray}{72.2} &41.9 &53.9 &62.2 &65.5 &\textcolor{darkgray}{72.2} \\
\ \pretrain CSC~\cite{hou2021csc} &28.9 &49.8 &59.4 &64.6 &\textcolor{darkgray}{73.8} &55.5 &60.5 &65.9 &68.2 &\textcolor{darkgray}{73.8} \\
\ \pretrain MSC~\cite{wu2023msc} &29.2 &50.7 &61.0 &64.9 &\textcolor{darkgray}{75.4} &61.0 &65.6 &68.9 &69.6 &\textcolor{darkgray}{75.4} \\
\cmidrule(lr){1-11}
\scratch PTv3~\cite{wu2024ptv3} &25.8 &48.9 &61.0 &67.0 &\textcolor{darkgray}{77.2} &60.1 &67.9 &71.4 &72.7 &\textcolor{darkgray}{77.2} \\
\ \pretrain PPT~\cite{wu2024ppt}~(sup.) &31.1 &52.6 &63.3 &68.2 &\textcolor{darkgray}{78.2} &62.4 &69.1 &74.3 &75.5 &\textcolor{darkgray}{78.2} \\
\ \pretrain Sonata~\cite{wu2025sonata} (lin.) &43.6 &62.5 &68.6 &69.8 &\textcolor{darkgray}{72.5} &69.0 &70.5 &71.1 &71.5 &\textcolor{darkgray}{72.5} \\
\ \pretrain Sonata~\cite{wu2025sonata} (dec.) &44.5 &64.1 &69.8 &72.5 &\textcolor{darkgray}{79.1} &69.8 &73.1 &75.0 &76.3 &\textcolor{darkgray}{79.1} \\
\ \pretrain Sonata~\cite{wu2025sonata} (f.t.) &45.3 &65.7 &72.4 &72.8 &\textcolor{darkgray}{79.4} &70.5 &73.6 &76.0 &77.0 &\textcolor{darkgray}{79.4} \\
\cellcolor[HTML]{f3f7fc}\ \pretrain Concerto (lin.) &\cellcolor[HTML]{def3e6}\textbf{48.2} &\cellcolor[HTML]{def3e6}\textbf{69.1} &73.6 &\cellcolor[HTML]{eef9f2}75.0 &\textcolor{darkgray}{77.3} &\cellcolor[HTML]{def3e6}\textbf{73.9} &\cellcolor[HTML]{eef9f2}75.2 &76.2 &76.3 &\textcolor{darkgray}{77.3} \\
\cellcolor[HTML]{f3f7fc}\ \pretrain Concerto (dec.) &44.6 &67.9 &\cellcolor[HTML]{eef9f2}73.7 &74.6 &\textcolor{darkgray}{79.5} &72.6 &74.6 &\cellcolor[HTML]{eef9f2}76.7 &\cellcolor[HTML]{eef9f2}77.6 &\textcolor{darkgray}{79.5} \\
\cellcolor[HTML]{f3f7fc}\ \pretrain Concerto (f.t.) &\cellcolor[HTML]{eef9f2}46.5 &\cellcolor[HTML]{eef9f2}69.0 &\cellcolor[HTML]{def3e6}\textbf{75.3} &\cellcolor[HTML]{def3e6}\textbf{76.1} &\textcolor{darkgray}{80.7} &\cellcolor[HTML]{eef9f2}73.3 &\cellcolor[HTML]{def3e6}\textbf{76.7} &\cellcolor[HTML]{def3e6}\textbf{77.6} &\cellcolor[HTML]{def3e6}\textbf{78.4} &\textcolor{darkgray}{80.7} \\
\bottomrule
\end{tabular}

        \vspace{-2mm}
        \subcaption{Data efficiency. We adopt the ScanNet Data Efficient~\cite{hou2021csc} benchmark and compare the validation mIoU(\%) results of Concerto with previous methods in three evaluation protocols.}
        \label{subtab:semseg_data_efficiency}
        \vspace{-2mm}
    \end{minipage}
    \caption{\textbf{Semantic segmentation.} We train Concerto on ScanNet~\cite{dai2017scannet}, ScanNet++~\cite{yeshwanth2023scannet++}, Structured3D~\cite{zheng2020structured3d}, S3DIS~\cite{armeni2016s3dis}, ArkitScenes~\cite{dehghan2021arkitscenes}, and HM3D~\cite{ramakrishnan2021hm3d} datasets, utilizing ScanNet, ScanNet200, ScanNet++, and S3DIS to evaluate the model by linear probing, decoder probing, and full fine-tuning and ScanNet Data Efficient~\cite{hou2021csc} to evaluate the data efficiency. The pre-training setting is the default, described in \tabref{tab:ablation}. More specific pre-training details are available in the Appendix.}
    \label{tab:semseg}
    \vspace{-10mm}
\end{table*}

\vspace{-1mm}
\mypara{Semantic segmentation.} In \tabref{subtab:semseg_fine_tuning}, we compared Concerto with the previous 3D encoder Sonata~\cite{wu2025sonata} on semantic segmentation tasks across multiple datasets: ScanNet~\cite{dai2017scannet}, ScanNet200~\cite{rozenberszki2022scannet200}, ScanNet++~\cite{yeshwanth2023scannet++}, and S3DIS Area5~\cite{armeni2016s3dis} with full fine-tuning. Across all datasets, Concerto achieves SOTA performance with notable mIoU results, including 80.7\% on ScanNet, 39.2\% on ScanNet200, and 50.7\% on ScanNet++. The most significant improvement is seen in the ScanNet200 dataset, which contains 200 class categories. This suggests that while detecting fine-grained objects in sparse point clouds remains challenging, joint 2D-3D cross-modal learning enables Concerto to capture detailed semantic and geometry information, thus improving the model’s ability for such objects.

\vspace{-1mm}
\mypara{Instance segmentation.} In \tabref{tab:insseg}, we further validate the robustness of Concerto across 4 widely recognized instance segmentation benchmarks. Concerto demonstrates the strongest performance in all evaluation methods. Notably, decoder probing on ScanNet outperforms full fine-tuning, suggesting that Concerto learns rich, generalizable representations during pretraining without task-specific adjustments. This demonstrates the advantage of leveraging general pretrained representations, which reduces the risk of distorting pretrained representations and overfitting in fine-tuning.

\vspace{-1mm}
\mypara{Parameter efficiency.} In \tabref{subtab:semseg_param_efficiency}, we demonstrate Concerto's parameter efficiency using the simplest linear probing and decoder probing across 4 semantic segmentation benchmarks. In particular, Concerto outperforms supervised learning using the PTv3 backbone~\cite{wu2024ptv3} on all benchmarks with decoder probing. Even with linear probing, Concerto surpasses the supervised PTv3 on ScanNet200~\cite{rozenberszki2022scannet200} and S3DIS~\cite{armeni2016s3dis}. Compared to Sonata~\cite{wu2025sonata}, Concerto shows significant improvements on ScanNet200~\cite{rozenberszki2022scannet200} and ScanNet++~\cite{yeshwanth2023scannet++} with linear probing (+8.1\% and +6.7\% respectively). These results highlight a substantial improvement in scenes with larger numbers of classes.

\vspace{-1mm}
\mypara{Data efficiency.} In \tabref{subtab:semseg_data_efficiency}, we examine the data efficiency performance of Concerto on ScanNet Efficient Datasets~\cite{hou2021csc} with limited scenes and annotations. Concerto outperforms Sonata~\cite{wu2025sonata} across all evaluation protocols. Notably, linear probing results surpass decoder probing and even full fine-tuning (SFT) in extreme data-limited scenarios (1\%, 5\% limited scenes, and 20-point annotation per scene). This observation aligns with findings in the image domain~\cite{zhang2022fine}, where linear probing outperforms full fine-tuning in out-of-distribution situations. In our case, when training on limited data, the whole evaluation dataset becomes an out-of-distribution situation. This significant emerging property reveals two key insights: more generalizable representations and more efficient adaptation potential. This could signal a potential shift toward Low-Rank Adaptation (LoRA) methods~\cite{hu2022lora} for fine-tuning point cloud backbones. Detailed LoRA fine-tuning results are provided in the Appendix.

\begin{figure}
    \centering
    \includegraphics[width=1.0\textwidth]{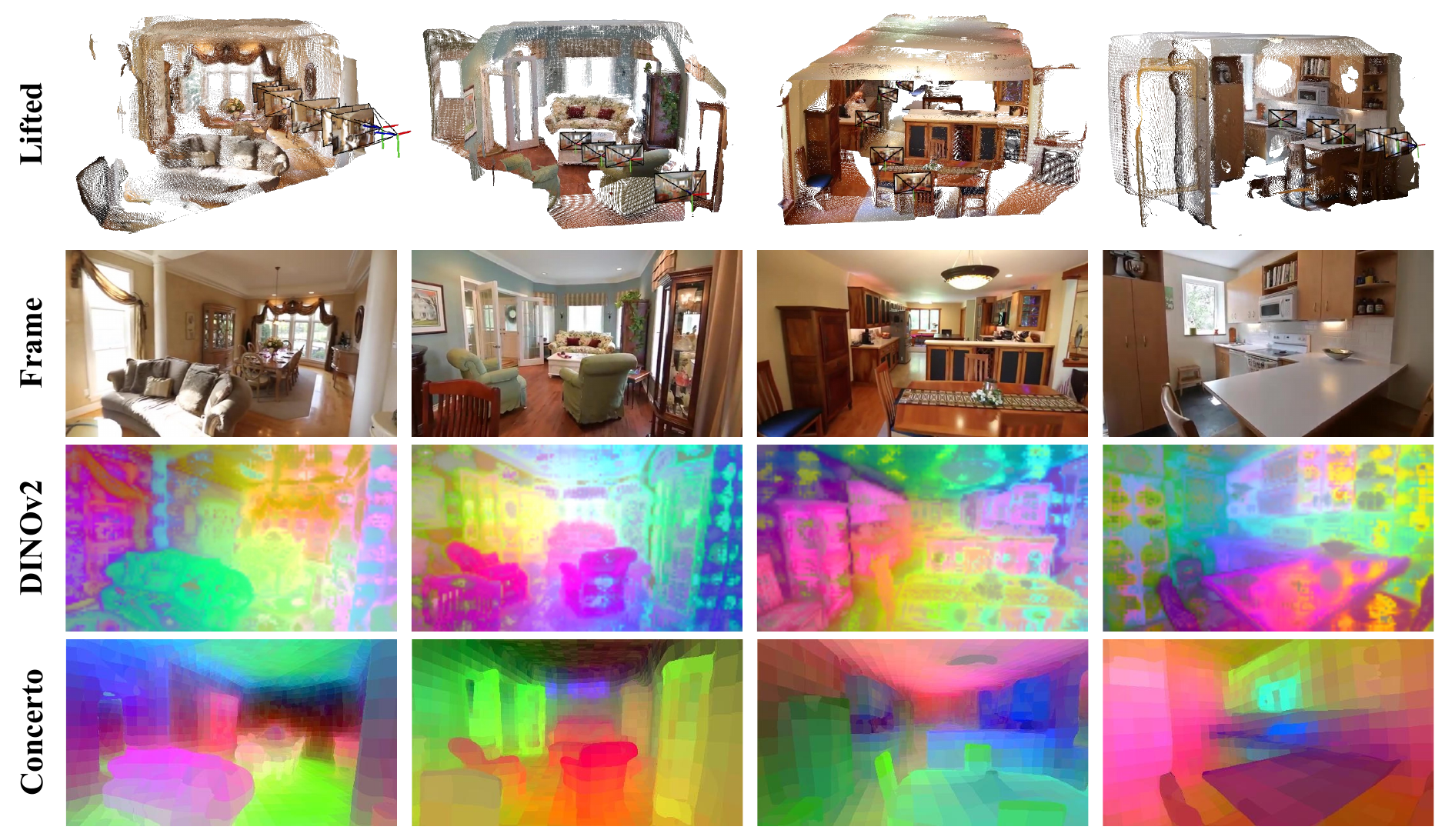}
    \vspace{-6mm}
    \caption{\textbf{Video spatial perception.} Concerto can be directly applied to video-lifted data (top row). The PCA visualizations (bottom two rows) illustrate that Concerto learns more fine-grained and semantically consistent features compared to DINOv2.}
    \vspace{-2mm}
    \label{fig:video}
\end{figure}

\mypara{Video processing.} As shown in the \figref{fig:video}, the variation of Concerto demonstrates strong adaptability to video-lifted data. We hypothesize that certain spatial-specific information is more effectively captured in the lifted space. By leveraging the current feed-forward reconstruction method VGGT~\cite{wang2025vggt} to reconstruct the point clouds from videos, we generate a diverse range of point cloud data. Incorporating these lifted point clouds into the pipeline allows Concerto to learn more generalizable representations, enhancing its ability for real-time video spatial perception. Moreover, by including video data into training datasets, we aim to extend the scaling ability of Concerto further. In \tabref{tab:scaling}, we provide results of different model sizes, where the large model variant trained with the additional video data demonstrates significant potential for further scaling. Details of this Concerto variation and the method of lifting are presented in the Appendix.

\begin{table*}[!t]
    \centering
    \tablestyle{3pt}{1.08}
    \begin{tabular}{lccccccccccccc}\toprule
Scale &\multicolumn{3}{c}{ScanNet Val} &\multicolumn{3}{c}{ScanNet200 Val} &\multicolumn{3}{c}{ScanNet++ Val} &\multicolumn{3}{c}{S3DIS Area 5} \\\cmidrule(lr){1-1} \cmidrule(lr){2-4} \cmidrule(lr){5-7} \cmidrule(lr){8-10} \cmidrule(lr){11-13}
Model Size &mIoU &mAcc &allAcc &mIoU &mAcc &allAcc &mIoU &mAcc &allAcc &mIoU &mAcc &allAcc \\
\midrule
5M(T) &67.7 &78.5 &87.4 &24.9 &34.4 &79.3 &33.7 &45.9 &82.7 &65.2 &73.6 &88.6 \\
39M(S) &76.6 &86.6 &91.5 &34.4 &46.3 &83.1 &43.1 &57.6 &86.2 &71.3 &80.4 &90.1 \\
108M(B) &77.3 &86.6 &91.7 &37.4 &49.5 &83.3 &45.6 &60.5 &86.5 &73.5 &81.3 &90.9 \\
207M(L) &77.5 &86.6 &92.1 &38.6 &49.8 &83.9 &46.3 &59.9 &86.7 &73.7 &81.4 &91.1 \\
\bottomrule
\end{tabular}

    \caption{\textbf{Scaling Up.} Model T, S, B is trained on point cloud datasets, while Model L is trained on point cloud datasets and an additional video dataset.}
    \label{tab:scaling}
\end{table*}

\begin{table*}[!t]
    \centering
    \tablestyle{0.5pt}{1.08}
    \begin{tabular}{y{22.5mm}rz{8mm}x{8mm}x{8mm}x{8mm}x{8mm}x{8mm}x{8mm}x{8mm}x{8mm}x{8mm}x{8mm}x{8mm}x{8mm}}\toprule
Ins. Seg. &\multicolumn{2}{c}{Params} &\multicolumn{3}{c}{ScanNet Val~\cite{dai2017scannet}} &\multicolumn{3}{c}{ScanNet200 Val~\cite{rozenberszki2022scannet200}} &\multicolumn{3}{c}{ScanNet++ Val~\cite{yeshwanth2023scannet++}} &\multicolumn{3}{c}{S3DIS Area 5~\cite{armeni2016s3dis}} \\
\cmidrule(lr){1-1} \cmidrule(lr){2-3} \cmidrule(lr){4-6} \cmidrule(lr){7-9} \cmidrule(lr){10-12} \cmidrule(lr){13-15}
Methods &Learn. &Pct. &$\text{mAP}_{25}$ & $\text{mAP}_{50}$ &mAP &$\text{mAP}_{25}$ & $\text{mAP}_{50}$ &mAP &$\text{mAP}_{25}$ & $\text{mAP}_{50}$ &mAP &$\text{mAP}_{25}$ & $\text{mAP}_{50}$ &mAP \\\midrule
\scratch \textcolor{darkgray}{PTv3}~\cite{wu2024ptv3} &\textcolor{darkgray}{124.8M} &\textcolor{darkgray}{100\%} &\textcolor{darkgray}{77.5} &\textcolor{darkgray}{61.7} &\textcolor{darkgray}{40.9} &\textcolor{darkgray}{40.1} &\textcolor{darkgray}{33.2} &\textcolor{darkgray}{23.1} &\textcolor{darkgray}{46.3} &\textcolor{darkgray}{39.6} &\textcolor{darkgray}{25.4} &\textcolor{darkgray}{55.7} &\textcolor{darkgray}{49.4} &\textcolor{darkgray}{37.8} \\
\ \pretrain MSC~\cite{wu2023msc} (lin.) &\tinyless0.2M &\tinyless0.2\% &13.3 &5.3 &2.3 &2.3 &1.0 &0.4 &4.8 &2.6 &1.3 &19.0 &13.0 &9.7 \\
\ \pretrain Sonata~\cite{wu2025sonata} (lin.) &\tinyless0.2M &\tinyless0.2\% &\cellcolor[HTML]{eef9f2}72.6 &\cellcolor[HTML]{eef9f2}53.9 &\cellcolor[HTML]{eef9f2}30.7 &\cellcolor[HTML]{eef9f2}30.9 &\cellcolor[HTML]{eef9f2}21.3 &\cellcolor[HTML]{eef9f2}10.9 &\cellcolor[HTML]{eef9f2}34.6 &\cellcolor[HTML]{eef9f2}26.5 &\cellcolor[HTML]{eef9f2}14.8 &\cellcolor[HTML]{eef9f2}45.8 &\cellcolor[HTML]{eef9f2}36.6 &\cellcolor[HTML]{def3e6}\textbf{26.1} \\
\cellcolor[HTML]{f3f7fc}\ \pretrain Concerto (lin.) &\tinyless0.2M &\tinyless0.2\% &\cellcolor[HTML]{def3e6}\textbf{75.4} &\cellcolor[HTML]{def3e6}\textbf{55.7} &\cellcolor[HTML]{def3e6}\textbf{31.1} &\cellcolor[HTML]{def3e6}\textbf{38.2} &\cellcolor[HTML]{def3e6}\textbf{27.7} &\cellcolor[HTML]{def3e6}\textbf{14.9} &\cellcolor[HTML]{def3e6}\textbf{40.1} &\cellcolor[HTML]{def3e6}\textbf{31.3} &\cellcolor[HTML]{def3e6}\textbf{18.2} &\cellcolor[HTML]{def3e6}\textbf{52.5} &\cellcolor[HTML]{def3e6}\textbf{40.8} &\cellcolor[HTML]{eef9f2}25.7 \\
\cmidrule(lr){1-15}
\ \pretrain Sonata~\cite{wu2025sonata} (dec.) &16.3M &13\% &76.8 &62.8 &40.8 &40.8 &33.3 &22.8 &39.4 &33.5 &22.7 &63.7 &57.1 &45.1 \\
\cellcolor[HTML]{f3f7fc}\ \pretrain Concerto (dec.) &16.3M &13\% &\cellcolor[HTML]{def3e6}\textbf{81.1} &\cellcolor[HTML]{def3e6}\textbf{64.2} &\cellcolor[HTML]{def3e6}\textbf{42.7} &\cellcolor[HTML]{def3e6}\textbf{41.8} &\cellcolor[HTML]{def3e6}\textbf{34.4} &\cellcolor[HTML]{def3e6}\textbf{24.0} &\cellcolor[HTML]{def3e6}\textbf{42.2} &\cellcolor[HTML]{def3e6}\textbf{35.3} &\cellcolor[HTML]{def3e6}\textbf{23.4} &\cellcolor[HTML]{def3e6}\textbf{66.8} &\cellcolor[HTML]{def3e6}\textbf{58.1} &\cellcolor[HTML]{def3e6}\textbf{45.1} \\
\cmidrule(lr){1-15}
\ \pretrain MSC~\cite{wu2023msc} (f.t.) &124.8M &100\% &78.4 &62.9 &41.1 &40.5 &33.8 &23.4 &- &- &- &56.3 &50.5 &38.1 \\
\ \pretrain PPT~\cite{wu2024ppt}~(sup.) &124.8M &100\% &78.9 &63.5 &42.1 &40.8 &34.1 &24.0 &- &- &- &57.5 &51.2 &39.7 \\
\ \ \pretrain Sonata~\cite{wu2025sonata} (f.t.) &124.8M &100\% &\cellcolor[HTML]{eef9f0}79.2 &\cellcolor[HTML]{eef9f0}63.9 &\cellcolor[HTML]{eef9f0}42.4 &\cellcolor[HTML]{eef9f0}42.1 &\cellcolor[HTML]{eef9f0}35.6 &\cellcolor[HTML]{eef9f0}25.4 &\cellcolor[HTML]{eef9f0}43.3 &\cellcolor[HTML]{eef9f0}36.5 &\cellcolor[HTML]{eef9f0}24.6 &\cellcolor[HTML]{eef9f2}63.8 &\cellcolor[HTML]{eef9f2}57.4 &\cellcolor[HTML]{eef9f2}45.5 \\
\cellcolor[HTML]{f3f7fc}\ \pretrain Concerto (f.t.) &124.8M &100\% &\cellcolor[HTML]{def3e4}\textbf{79.5} &\cellcolor[HTML]{def3e4}\textbf{64.9} &\cellcolor[HTML]{def3e4}\textbf{42.9} &\cellcolor[HTML]{def3e4}\textbf{45.8} &\cellcolor[HTML]{def3e4}\textbf{38.7} &\cellcolor[HTML]{def3e4}\textbf{27.4} &\cellcolor[HTML]{def3e4}\textbf{44.3} &\cellcolor[HTML]{def3e4}\textbf{38.3} &\cellcolor[HTML]{def3e4}\textbf{26.0} &\cellcolor[HTML]{def3e6}\textbf{67.5} &\cellcolor[HTML]{def3e6}\textbf{61.0} &\cellcolor[HTML]{def3e6}\textbf{46.4} \\
\bottomrule
\end{tabular}

    \vspace{-2mm}
    \caption{\textbf{Instance segmentation.} Concerto demonstrates the strongest performance for instance segmentation across 4 datasets with all evaluation protocols.}
    \label{tab:insseg}
    \vspace{-2mm}
\end{table*}

\begin{wraptable}{r}{0.4\textwidth}
  \centering
    \vspace{1mm}
    \tablestyle{5pt}{1.08}
    \begin{tabular}{lrrrr}\toprule
Language &\multicolumn{3}{c}{ScanNet Val} \\
\cmidrule(lr){1-1}\cmidrule(lr){2-4}
Methods &mIoU &mAcc &allAcc \\\midrule
MSC~\cite{wu2023msc} &12.42 &18.55 &59.89 \\
Sonata~\cite{wu2025sonata} &\cellcolor[HTML]{eef9f2}41.71 &\cellcolor[HTML]{eef9f2}61.29 &\cellcolor[HTML]{eef9f2}78.86 \\
\cellcolor[HTML]{f3f7fc}Concerto &\cellcolor[HTML]{def3e4}44.56 &\cellcolor[HTML]{def3e4}64.76 &\cellcolor[HTML]{def3e4}80.76 \\
\bottomrule
\end{tabular}

    \vspace{-2mm}
    \caption{\textbf{Language probing.}}
    \label{tab:lanlin}
    \vspace{-4mm}
\end{wraptable}

\mypara{Language probing.} Language develops from a fundamental understanding of the physical world, which motivates many works on language alignment with 3D knowledge as ~\cite{conceptfusion, huang2024openins3d}. In \tabref{tab:lanlin}, we demonstrate Concerto's ability to formulate concepts similar to human language, paving the way for future exploration of alignment with text-based semantic spaces. With linear probing trained on the same datasets as the pretraining stage, we translate Concerto's representations to language space by aligning LSeg~\cite{li2022lseg} image encoder output to our linear probing output. Without ground truth labels, Concerto achieves 44.56\% mIoU on ScanNet zero-shot segmentation. Although this lags behind the 77.3\% mIoU by supervised linear probing, we expect that further language-conditioned probing will yield comparable results, marking a significant step toward bridging 3D spatial representations with text.

\begin{table*}[!t]
    \begin{minipage}{0.415\textwidth}
    \centering
        \tablestyle{1pt}{0.9}
        \begin{tabular}{x{10mm}|x{10mm}x{10mm}x{10mm}x{10mm}}
cross &o. c. &\cellcolor[HTML]{f3f7fc}c. s. &o. c. &c. s. \\
intra &o. c. &\cellcolor[HTML]{f3f7fc}o. c. &c. s. &c. s. \\\midrule
lin. &60.7 &\cellcolor[HTML]{eef9f2}\textbf{75.6} &31.6 &74.7 \\
dec. &72.0 &\cellcolor[HTML]{eef9f2}\textbf{78.6} &66.0 &78.3 \\
\end{tabular}

        \vspace{-1mm}
        \subcaption{\textbf{Criteria type.}
        o.c.: online clustering criteria in DINOv2~\cite{oquab2023dinov2}; c.s.: cosine similarity.
        }
        \label{subtab:ablation_criteria_type}
        \vspace{1mm}
    \end{minipage}
    \hspace{0.06\textwidth}
    \begin{minipage}{0.495\textwidth}
    \centering
        \tablestyle{1pt}{1.13}
        \begin{tabular}{x{15mm}|x{10mm}x{10mm}x{10mm}x{10mm}x{10mm}}
image usage&0\% &20\% &50\% &70\% &\cellcolor[HTML]{f3f7fc}100\% \\\midrule
lin. &70.9 &73.5 &75.2 &75.4 &\cellcolor[HTML]{eef9f0}\textbf{75.6} \\
dec. &77.4 &77.4 &77.8 &78.4 &\cellcolor[HTML]{eef9f2}\textbf{78.6} \\
\end{tabular}

        \vspace{-1mm}
        \subcaption{\textbf{Image usage.}
        We control the image usage ratio to present the effectiveness of joint cross-modal learning.
        }
        \label{subtab:ablation_image_usage}
        \vspace{1mm}
    \end{minipage} \\
    \begin{minipage}{0.37\textwidth}
    \centering
        \tablestyle{1pt}{1.08}
        \begin{tabular}{x{20mm}|x{10mm}x{10mm}x{10mm}}
criteria weight &\cellcolor[HTML]{f3f7fa}1:2 &2:2 &3:2 \\\midrule
lin. &75.6 &\cellcolor[HTML]{eef9f2}\textbf{76.1} &75.7 \\
dec. &78.6 &78.6 &\cellcolor[HTML]{eef9f2}\textbf{78.8} \\
\end{tabular}

        \vspace{-1mm}
        \subcaption{\textbf{Criteria weight.} A suitable weight ratio (cross:intra) is needed to balance intra- and cross-modal components.}
        \label{subtab:ablation_criteria_weight}
        \vspace{1mm}
    \end{minipage}
    \hspace{0.03\textwidth}
    \begin{minipage}{0.26\textwidth}
    \centering
        \tablestyle{1pt}{1.08}
        \begin{tabular}{x{15mm}|x{10mm}x{10mm}x{10mm}}
img. aug. &w &\cellcolor[HTML]{f3f7fc}w/o \\\midrule
lin. &\cellcolor[HTML]{eef9f2}\textbf{76.7} &75.6 \\
dec. &\cellcolor[HTML]{eef9f2}\textbf{78.6} &78.6 \\
\end{tabular}

        \vspace{-1mm}
        \subcaption{\textbf{Image augmentation.} Weak image augmentation shows a positive impact.
        }
        \label{subtab:ablation_image_augmentation}
        \vspace{1mm}
    \end{minipage}
        \hspace{0.03\textwidth}
    \begin{minipage}{0.26\textwidth}
    \centering
        \tablestyle{1pt}{1.08}
        \begin{tabular}{x{15mm}|x{10mm}x{10mm}x{10mm}}
vis. points&\cellcolor[HTML]{f3f7fa}$\times$1 &$\times$2 \\\midrule
lin. &\cellcolor[HTML]{eef9f2}\textbf{75.6} &75.5 \\
dec. &\cellcolor[HTML]{eef9f2}\textbf{78.6} &78.1 \\
\end{tabular}

        \vspace{-1mm}
        \subcaption{\textbf{Visible points}. Fewer visible points, better performance. $\times$1: 65536 points.}
        \label{subtab:ablation_visiable_points}
        \vspace{1mm}
    \end{minipage} \\
    \begin{minipage}{0.37\textwidth}
    \centering
        \tablestyle{1pt}{1.08}
        \begin{tabular}{x{20mm}|x{10mm}x{10mm}x{10mm}x{10mm}}
upcast level &2 &\cellcolor[HTML]{f3f7fa}3 &4 \\\midrule
lin. &75.1 &\cellcolor[HTML]{eef9f2}\textbf{75.6} &75.3 \\
dec. &78.5 &\cellcolor[HTML]{eef9f2}\textbf{78.6} &78.2 \\
\end{tabular}

        \vspace{-1mm}
        \subcaption{\textbf{Upcast level.} Upcast level, processed as~\cite{wu2025sonata}, refers to the concatenation level of features in cross-modal learning here.}
        \label{subtab:ablation_upcast}
    \end{minipage}
    \hspace{0.03\textwidth}
    \begin{minipage}{0.26\textwidth}
    \centering
        \tablestyle{1pt}{1.08}
        \begin{tabular}{x{15mm}|x{10mm}x{10mm}x{10mm}}
data scale &\cellcolor[HTML]{f3f7fc}23k &40k \\\midrule
lin. &75.6 &\cellcolor[HTML]{eef9f0}\textbf{76.6} \\
dec. &78.6 &\cellcolor[HTML]{eef9f0}\textbf{79.2} \\
\end{tabular}

        \vspace{-1mm}
        \subcaption{\textbf{Data scale.}
        23k is the default ablation setting and the total dataset contains 40k.}
        \label{subtab:ablation_data_scale}
    \end{minipage}
        \hspace{0.03\textwidth}
    \begin{minipage}{0.26\textwidth}
    \centering
        \tablestyle{1pt}{1.08}
        \begin{tabular}{x{15mm}|x{10mm}x{10mm}x{10mm}}
model scale &\cellcolor[HTML]{f3f7fc}s &b \\\midrule
lin. &76.6 &\cellcolor[HTML]{eef9f0}\textbf{77.3} \\
dec. &79.2 &\cellcolor[HTML]{eef9f0}\textbf{79.5} \\
\end{tabular}

        \vspace{-1mm}
        \subcaption{\textbf{Model scale.}
        Training on 40k. s: 39M backbone; b: 108M backbone.
        }
        \label{subtab:ablation_model_scale}
    \end{minipage} \\
    \caption{\textbf{Ablation study.} The default ablation setup trains on ScanNet~\cite{dai2017scannet} and Structured3d~\cite{zheng2020structured3d} with 39M PTv3~\cite{wu2024ptv3} model as in Sonata~\cite{wu2025sonata}. If not specified, other default settings are in the Appendix. For \tabref{subtab:ablation_data_scale} and \tabref{subtab:ablation_model_scale}, we scale the setup to match the model used in the main results. All of our designs are enabled by default. Default settings are marked in \colorbox[HTML]{f3f7fc}{blue}.}
    \label{tab:ablation}
    \vspace{-5mm}
\end{table*}

\vspace{-1mm}
\subsection{Ablation Study}
\vspace{-2mm}

In this section, we ablate intriguing properties for Concerto in \tabref{tab:ablation} with the default setting in captions, evaluating the ablation on ScanNet semantic segmentation using linear and decoder probing.

\vspace{-1mm}
\mypara{Joint cross-modal learning.} In \tabref{tab:2dx3d}, we investigate the influence of joint cross-modal learning by comparing Concerto with strong baseline models DINOv2 and Sonata. Concerto outperforms both and surpasses their native feature concatenation. These results demonstrate that joint cross-modal learning does more than merely merge information from different modalities; it enables the model to learn richer emerging representations that were previously unattainable.

\vspace{-1mm}
\mypara{Criteria type.} In \tabref{subtab:ablation_criteria_type}, we show that using cosine similarity as the loss function in the cross-modal joint embedding prediction component and cross-entropy-based online clustering loss from DINOv2~\cite{oquab2023dinov2} in the self-distillation component facilitates joint 2D-3D self-supervised learning in latent space. This combination reduces strict constraints and minimizes conflicts between cross-modal and intra-modal learning, enabling a smoother joint learning process for the two objectives.

\vspace{-1mm}
\mypara{Image usage.} In \tabref{subtab:ablation_image_usage}, we further investigate the effect of multisensory interactions by varying the input ratio of point clouds with images. The results show that even with a small ratio, such as 20\%, joint cross-modal learning is effective, leading to improvements in linear probing. When the image usage ratio reaches 50\%, the linear probing result is comparable to that with 100\% image usage, while decoder probing continues to show potential for further improvement. These results suggest that shallow linear representations are easier to discover with a smaller proportion of images, while deeper representations, which require decoder probing, benefit from a higher image usage ratio.

\vspace{-1mm}
\mypara{Criteria weight.} In \tabref{subtab:ablation_criteria_weight}, we observe that the criteria weight ratio between cross-modal and intra-modal components affects performance. Given the distinct objectives of intra-modal and cross-modal learning, maintaining a balanced loss-weight ratio is essential for optimal performance. The 2:2 ratio outperforms the others in linear probing.

\vspace{-1mm}
\mypara{Image augmentation.} Data augmentations are crucial in self-supervised learning. As Sonata~\cite{wu2025sonata} already explores point cloud augmentations in self-distillation, we focus on the cross-modal image augmentations here. Initially, we follow DINOv2 strong data augmentations, which results in a lower performance with a linear probing mIoU of 75.27\%. However, when we apply less aggressive augmentations, the performance surpasses our default setting without image augmentations, as in \tabref{subtab:ablation_image_augmentation}. The details of the image augmentations are provided in the Appendix. Since the image encoder here is frozen, it does not benefit from data augmentations. Additionally, overly aggressive image augmentation may confuse the point encoder with excessive distortions. Thus, careful selection of image augmentations is essential. Currently, our model does not apply image augmentations, and we plan to explore this in future updates.

\vspace{-1mm}
\mypara{Visible points.} In \tabref{subtab:ablation_visiable_points}, we investigate the impact of visible points in the point cloud from image views. We hypothesize that while a large amount of matched point-pixel pairs can offer more complete information, the smaller number of pairs forces the model to predict across modalities for the surrounding context. As the task becomes more challenging, the model is encouraged to dig deeper into semantics, leading to better performance. In ablation, the performance is quite similar as the point numbers we selected are still quite small, compared with the number of image pixels and points in point clouds.

\vspace{-1mm}
\mypara{Upcast level.} While Sonata~\cite{wu2025sonata} studies the influence of upcast levels on self-supervised learning representations, we present the performance of different cross-modal feature upcast levels in \tabref{subtab:ablation_upcast}. The model at upcast level 3 achieves the best performance, indicating that the upcast level 3 is close to the corresponding scales of the image and point cloud. Additionally, level 3 outperforms level 4 as level 4 may retain too many low-level details that are not beneficial for joint embedding learning. Furthermore, level 3 surpasses level 2, as the use of Sonata's self-distillation technique at level 2 might introduce conflicts between intra-modal self-supervised learning and cross-modal joint learning of the same upcast level, ultimately leading to negative effects.

\vspace{-1mm}
\mypara{Data scale and model weight.} Additionally, following the approach of Sonata~\cite{wu2025sonata}, we scale our datasets from 23k to 40k, resulting in a significant improvement shown in \tabref{subtab:ablation_data_scale}. Likewise, aligning our model size with that of Sonata (108M) further enhances the performance, as demonstrated in \tabref{subtab:ablation_model_scale}. Larger datasets provide more diverse and comprehensive information, enabling the model to learn more general patterns. As the model size increases, it becomes better equipped to capture complex relationships. Combining both naturally leads to more generalized representation outputs.

\vspace{-2mm}
\section{Related Work}
\vspace{-2mm}

\mypara{2D image self-supervised learning.} Aimed at utilizing oceans of unlabeled data, image self-supervised learning has seen significant progress~\cite{zhang2022dino,oquab2023dinov2,assran2023jepa,he2020moco,chen2020simclr}. These methods always focus on learning invariant representations through transformations or augmentations of the data. One of the most notable achievements in this field is DINOv2~\cite{oquab2023dinov2}, producing high-quality image representations. Building on the success of DINOv2, Concerto extends its reliable 2D image representations to the cross-modal domain, incorporating both 2D image and 3D point cloud data for superior representations in the 3D domain.

\vspace{-1mm}
\mypara{3D point self-supervised learning.} While self-supervised learning has made significant progress in the image domain, it is still in the starting stage of 3D point clouds. Building on the success of Sonata~\cite{wu2025sonata}, we further extend the previous works~\cite{xie2020pointcontrast, wang2024gc, wu2023msc} on unimodal self-supervised learning with scene-level data to joint 2D-3D self-supervised learning for superior representation extracting ability. Before Sonata~\cite{wu2025sonata}, most of the point self-supervised learning works suffer from geometry shortcuts due to the sparse and unordered nature of point clouds. Based on its predecessor, Concerto includes 2D images in its pipeline: leveraging point clouds to predict image features from DINOv2~\cite{oquab2023dinov2} by cross-modal joint embedding prediction and including 200k video-lifted point clouds by feed-forward reconstruction methods~\cite{wang2025vggt} into training datasets.

\vspace{-1mm}
\mypara{Spatial understanding with joint 2D-3D data.} With the rapid advancements in 2D self-supervised learning and its remarkable performance, many methods for point cloud representation now incorporate image features into their pipelines. Approaches such as lifting projections~\cite{t2024lift3d, fan2024lsm}, differentiable rendering~\cite{yue2024fit3d, kobayashi2022decomposing, zhang2024condense}, direct distillation ~\cite{chen2023clip2scene, zhu2023ponderv2, zeid2025dinoinroom, chen2023bridge3d}, attention-based feature fusion ~\cite{Zhang_2025}, and using text-aligned image encoders for open-vocabulary tasks~\cite{peng2023openscene3d, takmaz2023openmask3d, wang2024ggsd, jain2025univlg} aim to incorporating image features in 3D learning. However, these methods primarily focus on imitating image features in point cloud representations, often overlooking the full potential of multi-modal interaction. Recently, Locate3D~\cite{arnaud2025locate3d} seeks to develop generalizable representations beyond 2D image features but still relies on 2D image features during inference with a complicated pipeline. In contrast, Concerto utilizes joint 2D-3D embedding prediction during training, resulting in unified and rich representations beyond individual 2D or 3D features and their simple combination, enabling superior representations in inference with only point clouds.

\vspace{-1mm}
\section{Conclusion and Discussion}
\vspace{-2mm}

In this work, we present Concerto, achieving SOTA performance across multiple benchmarks. Additionally, we present the variation of Concerto for video spatial perception and the interlude of Concerto, which explores potential future alignment with text spaces. Concerto holds great promise for joint multi-modal self-supervised learning. Currently, it excels as a joint 2D-3D self-supervised learning model in the 3D domain, delivering superior performance in spatial representation learning. However, our goal extends beyond this. We discuss limitations and future works as follows:
\begin{itemize}
    \item \textit{Native multi-modal representation learning.} Our current training recipe freezes the image encoder, treating it as a static feature extractor. A compelling future direction is to unfreeze both the image and point cloud encoders for joint native multi-modal representation pre-training. This approach would enable the two modalities to mutually enhance one another during learning, fostering the development of a more robust shared representation space.
    \item \textit{Deep semantic grounding of language in point clouds.} While our work Concerto employs linear probing as an effective metric for evaluating feature quality, this method deliberately promotes a shallow alignment between point clouds and language to avoid the influence of post-training part on the evaluation performance of pretraining features. For real-world applications, this is insufficient to just have a shallow language alignment with specific key terms. A critical next step is to develop architectures and training objectives that move beyond simple feature alignment towards deep semantic grounding. The goal is to enable the learned representations to comprehend and respond to nuanced, indirect, or compositional linguistic descriptions, which remains a significant open challenge.
    \item \textit{Unified self-supervised learning paradigm for diverse point cloud domains.} Self-supervised learning for point clouds has historically been fragmented, with models tailored to specific domains (e.g., indoor, outdoor, object-level) to handle their distinct characteristics of scale and density. We believe that a unified pre-training paradigm trained on the data from different domains can produce more powerful and generalizable representations.By incorporating varied data sources like lidar point clouds, video-lifted point clouds, object-centric point clouds, and dynamic egocentric point clouds, a single self-supervised model can learn features that are robust to domain shifts. This enhanced generalization is expected to significantly boost performance on a wide array of downstream tasks, even those confined to a single domain.
\end{itemize}

\mypara{Acknowledgments.}
The research presented in this paper was supported by the National Natural Science Foundation of China (No. 62422606, 62201484).

{
\clearpage
\small
\renewcommand\UrlFont{\color{Gray}\ttfamily}
\bibliographystyle{ieeenat_fullname}
\bibliography{main}

@String(CVPR= {IEEE Conf. Comput. Vis. Pattern Recog.})

@String(ICCV= {Int. Conf. Comput. Vis.})

@String(ECCV= {Eur. Conf. Comput. Vis.})

@String(ICLR = {Int. Conf. Learn. Represent.})

@String(CVPR  = {CVPR})

@String(ICCV  = {ICCV})

@String(ECCV  = {ECCV})

@String(ICLR  = {ICLR})

@inproceedings{choy20194d,
  title     = {4d spatio-temporal convnets: Minkowski convolutional neural networks},
  author    = {Choy, Christopher and Gwak, JunYoung and Savarese, Silvio},
  booktitle = {CVPR},
  year      = {2019}
}

@inproceedings{dai2017scannet,
  title     = {ScanNet: Richly-annotated 3D Reconstructions of Indoor Scenes},
  author    = {Dai, Angela and Chang, Angel X. and Savva, Manolis and Halber, Maciej and Funkhouser, Thomas and Nie{\ss}ner, Matthias},
  booktitle = {CVPR},
  year      = {2017}
}

@inproceedings{armeni2016s3dis,
  title     = {3D Semantic Parsing of Large-Scale Indoor Spaces},
  author    = {Iro Armeni and Ozan Sener and Amir R. Zamir and Helen Jiang and Ioannis Brilakis and Martin Fischer and Silvio Savarese},
  booktitle = {CVPR},
  year      = {2016}
}

@article{zhang2022dino,
  title={DINO: DETR with Improved DeNoising Anchor Boxes for End-to-End Object Detection},
  author={Hao Zhang and Feng Li and Shilong Liu and Lei Zhang and Hang Su and Jun Zhu and Lionel M. Ni and Heung-Yeung Shum},
  journal={arXiv:2203.03605},
  year={2022}
}

@inproceedings{xie2020pointcontrast,
  title={Pointcontrast: Unsupervised pre-training for 3d point cloud understanding},
  author={Xie, Saining and Gu, Jiatao and Guo, Demi and Qi, Charles R and Guibas, Leonidas and Litany, Or},
  booktitle={ECCV},
  year={2020}
}

@inproceedings{xie2022simmim,
  title={Simmim: A simple framework for masked image modeling},
  author={Xie, Zhenda and Zhang, Zheng and Cao, Yue and Lin, Yutong and Bao, Jianmin and Yao, Zhuliang and Dai, Qi and Hu, Han},
  booktitle={CVPR},
  year={2022}
}

@inproceedings{dehghan2021arkitscenes,
title={{ARK}itScenes - A Diverse Real-World Dataset for 3D Indoor Scene Understanding Using Mobile {RGB}-D Data},
author={Gilad Baruch and Zhuoyuan Chen and Afshin Dehghan and Tal Dimry and Yuri Feigin and Peter Fu and Thomas Gebauer and Brandon Joffe and Daniel Kurz and Arik Schwartz and Elad Shulman},
booktitle={NeurIPSW},
year={2021}
}

@inproceedings{chen2020simclr,
  title={A simple framework for contrastive learning of visual representations},
  author={Chen, Ting and Kornblith, Simon and Norouzi, Mohammad and Hinton, Geoffrey},
  booktitle={ICML},
  year={2020},
}

@inproceedings{he2020moco,
  title={Momentum contrast for unsupervised visual representation learning},
  author={He, Kaiming and Fan, Haoqi and Wu, Yuxin and Xie, Saining and Girshick, Ross},
  booktitle={CVPR},
  year={2020}
}

@inproceedings{caron2020swav,
  title={Unsupervised learning of visual features by contrasting cluster assignments},
  author={Caron, Mathilde and Misra, Ishan and Mairal, Julien and Goyal, Priya and Bojanowski, Piotr and Joulin, Armand},
  booktitle={NeurIPS},
  year={2020}
}

@inproceedings{caron2021emerging,
  title={Emerging properties in self-supervised vision transformers},
  author={Caron, Mathilde and Touvron, Hugo and Misra, Ishan and J{\'e}gou, Herv{\'e} and Mairal, Julien and Bojanowski, Piotr and Joulin, Armand},
  booktitle={CVPR},
  year={2021}
}

@inproceedings{hou2021csc,
  title={Exploring data-efficient 3d scene understanding with contrastive scene contexts},
  author={Hou, Ji and Graham, Benjamin and Nie{\ss}ner, Matthias and Xie, Saining},
  booktitle={CVPR},
  year={2021}
}

@inproceedings{wu2022ptv2,
  title     = {Point transformer V2: Grouped Vector Attention and Partition-based Pooling},
  author    = {Wu, Xiaoyang and Lao, Yixing and Jiang, Li and Liu, Xihui and Zhao, Hengshuang},
  booktitle = {NeurIPS},
  year      = {2022}
}

@inproceedings{rozenberszki2022scannet200,
    title={Language-Grounded Indoor 3D Semantic Segmentation in the Wild},
    author={Rozenberszki, David and Litany, Or and Dai, Angela},
    booktitle={ECCV},
    year={2022}
}

@inproceedings{wu2023msc,
  title={Masked Scene Contrast: A Scalable Framework for Unsupervised 3D Representation Learning},
  author={Wu, Xiaoyang and Wen, Xin and Liu, Xihui and Zhao, Hengshuang},
  booktitle={CVPR},
  year={2023}
}

@inproceedings{zheng2020structured3d,
  title={Structured3d: A large photo-realistic dataset for structured 3d modeling},
  author={Zheng, Jia and Zhang, Junfei and Li, Jing and Tang, Rui and Gao, Shenghua and Zhou, Zihan},
  booktitle={ECCV},
  year={2020},
}

@inproceedings{caesar2020nuscenes,
  title={nuscenes: A multimodal dataset for autonomous driving},
  author={Caesar, Holger and Bankiti, Varun and Lang, Alex H and Vora, Sourabh and Liong, Venice Erin and Xu, Qiang and Krishnan, Anush and Pan, Yu and Baldan, Giancarlo and Beijbom, Oscar},
  booktitle={CVPR},
  year={2020}
}

@inproceedings{sun2020waymo,
  title={Scalability in perception for autonomous driving: Waymo open dataset},
  author={Sun, Pei and Kretzschmar, Henrik and Dotiwalla, Xerxes and Chouard, Aurelien and Patnaik, Vijaysai and Tsui, Paul and Guo, James and Zhou, Yin and Chai, Yuning and Caine, Benjamin and others},
  booktitle={CVPR},
  year={2020}
}

@article{zhu2023ponderv2,
  title={PonderV2: Pave the Way for 3D Foundataion Model with A Universal Pre-training Paradigm},
  author={Zhu, Haoyi and Yang, Honghui and Wu, Xiaoyang and Huang, Di and Zhang, Sha and He, Xianglong and He, Tong and Zhao, Hengshuang and Shen, Chunhua and Qiao, Yu and others},
  journal={arXiv:2310.08586},
  year={2023}
}

@inproceedings{wu2024ppt,
  title={Towards Large-scale 3D Representation Learning with Multi-dataset Point Prompt Training},
  author={Wu, Xiaoyang and Tian, Zhuotao and Wen, Xin and Peng, Bohao and Liu, Xihui and Yu, Kaicheng and Zhao, Hengshuang},
  booktitle={CVPR},
  year={2024}
}

@inproceedings{wu2024ptv3,
    title={Point Transformer V3: Simpler, Faster, Stronger},
    author={Wu, Xiaoyang and Jiang, Li and Wang, Peng-Shuai and Liu, Zhijian and Liu, Xihui and Qiao, Yu and Ouyang, Wanli and He, Tong and Zhao, Hengshuang},
    booktitle={CVPR},
    year={2024}
}

@inproceedings{wang2024gc,
  title={GroupContrast: Semantic-aware Self-supervised Representation Learning for 3D Understanding},
  author={Wang, Chengyao and Jiang, Li and Wu, Xiaoyang and Tian, Zhuotao and Peng, Bohao and Zhao, Hengshuang and Jia, Jiaya},
  booktitle={CVPR},
  year={2024}
}

@inproceedings{Pang2022pointmae,
  title     = {Masked autoencoders for point cloud self-supervised learning},
  author    = {Pang, Yatian and Wang, Wenxiao and Tay, Francis EH and Liu, Wei and Tian, Yonghong and Yuan, Li},
  booktitle = {ECCV},
  year      = {2022}
}

@inproceedings{yeshwanth2023scannet++,
  title={Scannet++: A high-fidelity dataset of 3d indoor scenes},
  author={Yeshwanth, Chandan and Liu, Yueh-Cheng and Nie{\ss}ner, Matthias and Dai, Angela},
  booktitle={ICCV},
  year={2023}
}

@article{oquab2023dinov2,
  title={DINOv2: Learning Robust Visual Features without Supervision},
  author={Oquab, Maxime and Darcet, Timothée and Moutakanni, Theo and Vo, Huy V. and Szafraniec, Marc and Khalidov, Vasil and Fernandez, Pierre and Haziza, Daniel and Massa, Francisco and El-Nouby, Alaaeldin and Howes, Russell and Huang, Po-Yao and Xu, Hu and Sharma, Vasu and Li, Shang-Wen and Galuba, Wojciech and Rabbat, Mike and Assran, Mido and Ballas, Nicolas and Synnaeve, Gabriel and Misra, Ishan and Jegou, Herve and Mairal, Julien and Labatut, Patrick and Joulin, Armand and Bojanowski, Piotr},
  journal={TMLR},
  year={2024}
}

@inproceedings{ramakrishnan2021hm3d,
  title={Habitat-Matterport 3D Dataset ({HM}3D): 1000 Large-scale 3D Environments for Embodied {AI}},
  author={Santhosh Kumar Ramakrishnan and Aaron Gokaslan and Erik Wijmans and Oleksandr Maksymets and Alexander Clegg and John M Turner and Eric Undersander and Wojciech Galuba and Andrew Westbury and Angel X Chang and Manolis Savva and Yili Zhao and Dhruv Batra},
  booktitle={NeurIPS},
  year={2021},
}

@inproceedings{avetisyan2024scenescript,
  title={SceneScript: Reconstructing Scenes With An Autoregressive Structured Language Model},
  author={Avetisyan, Armen and Xie, Christopher and Howard-Jenkins, Henry and Yang, Tsun-Yi and Aroudj, Samir and Patra, Suvam and Zhang, Fuyang and Frost, Duncan and Holland, Luke and Orme, Campbell and others},
  booktitle={ECCV},
  year={2024}
}

@inproceedings{assran2023jepa,
  title={Self-Supervised Learning from Images with a Joint-Embedding Predictive Architecture},
  author={Assran, Mahmoud and Duval, Quentin and Misra, Ishan and Bojanowski, Piotr and Vincent, Pascal and Rabbat, Michael and LeCun, Yann and Ballas, Nicolas},
  booktitle={CVPR},
  year={2023}
}

@inproceedings{yu2021pointbert,
  title={Point-BERT: Pre-Training 3D Point Cloud Transformers with Masked Point Modeling},
  author={Yu, Xumin and Tang, Lulu and Rao, Yongming and Huang, Tiejun and Zhou, Jie and Lu, Jiwen},
  booktitle={CVPR},
  year={2022}
}

@article{straub2024efm3d,
  title={EFM3D: A Benchmark for Measuring Progress Towards 3D Egocentric Foundation Models},
  author={Straub, Julian and DeTone, Daniel and Shen, Tianwei and Yang, Nan and Sweeney, Chris and Newcombe, Richard},
  journal={arXiv:2406.10224},
  year={2024}
}

@inproceedings{gu2023maniskill2,
  title={ManiSkill2: A Unified Benchmark for Generalizable Manipulation Skills},
  author={Gu, Jiayuan and Xiang, Fanbo and Li, Xuanlin and Ling, Zhan and Liu, Xiqiang and Mu, Tongzhou and Tang, Yihe and Tao, Stone and Wei, Xinyue and Yao, Yunchao and Yuan, Xiaodi and Xie, Pengwei and Huang, Zhiao and Chen, Rui and Su, Hao},
  booktitle={ICLR},
  year={2023}
}

@article{james2019rlbench,
  title={RLBench: The Robot Learning Benchmark \& Learning Environment},
  author={James, Stephen and Ma, Zicong and Rovick Arrojo, David and Davison, Andrew J.},
  journal={RAL},
  year={2020}
}

@inproceedings{zhang2022fine,
  title={Fine-tuning can distort pretrained features and underperform out-of-distribution},
  author={Zhang, Michael and Wang, Aditi Raghunathan and Sagawa, Shiori and Hashimoto, Tatsunori B and Liang, Percy},
  booktitle={ICLR},
  year={2022}
}

@inproceedings{wu2025sonata,
    title={Sonata: Self-Supervised Learning of Reliable Point Representations},
    author={Wu, Xiaoyang and DeTone, Daniel and Frost, Duncan and Shen, Tianwei and Xie, Chris and Yang, Nan and Engel, Jakob and Newcombe, Richard and Zhao, Hengshuang and Straub, Julian},
    booktitle={CVPR},
    year={2025}
}

@inproceedings{t2024lift3d,
  title     = {Lift3D: Zero-Shot Lifting of Any 2D Vision Model to 3D},
  author    = {T., Mukund Varma and Wang, Peihao and Fan, Zhiwen and Wang, Zhangyang and Su, Hao and Ramamoorthi, Ravi},
  booktitle = {CVPR},
  year      = {2024},
  eprint    = {https://arxiv.org/abs/2403.18922v1}
}

@inproceedings{yue2024fit3d,
  title     = {Improving 2D Feature Representations by 3D-Aware Fine-Tuning},
  author    = {Yue, Yuanwen and Das, Anurag and Engelmann, Francis and Tang, Siyu and Lenssen, Jan Eric},
  booktitle = {ECCV},
  year      = {2024},
  eprint    = {https://arxiv.org/abs/2407.20229v1}
}

@inproceedings{wang2024ggsd,
  title     = {Open Vocabulary 3D Scene Understanding via Geometry Guided Self-Distillation},
  author    = {Wang, Pengfei and Wang, Yuxi and Li, Shuai and Zhang, Zhaoxiang and Lei, Zhen and Zhang, Lei},
  booktitle = {ECCV},
  year      = {2024},
  eprint    = {https://arxiv.org/abs/2407.13362v1}
}

@inproceedings{peng2023openscene3d,
  title={Openscene: 3d scene understanding with open vocabularies},
  author={Peng, Songyou and Genova, Kyle and Jiang, Chiyu and Tagliasacchi, Andrea and Pollefeys, Marc and Funkhouser, Thomas and others},
  booktitle={CVPR},
  year={2023}
}

@inproceedings{chen2023bridge3d,
  title={Bridging the domain gap: Self-supervised 3d scene understanding with foundation models},
  author={Chen, Zhimin and Jing, Longlong and Li, Yingwei and Li, Bing},
  booktitle={NeurIPS},
  year={2023}
}

@inproceedings{zhang2024condense,
  title     = {{CONDENSE:} Consistent 2D/3D Pre-training for Dense and Sparse Features from Multi-View Images},
  author    = {Zhang, Xiaoshuai and Wang, Zhicheng and Zhou, Howard and Ghosh, Soham and Gnanapragasam, Danushen and Jampani, Varun and Su, Hao and Guibas, Leonidas J.},
  booktitle = {ECCV},
  year      = {2024},
  eprint    = {https://arxiv.org/abs/2408.17027v1}
}

@inproceedings{takmaz2023openmask3d,
  title     = {OpenMask3D: Open-Vocabulary 3D Instance Segmentation},
  author    = {Takmaz, Ay{\c{c}}a and Fedele, Elisabetta and Sumner, Robert W. and Pollefeys, Marc and Tombari, Federico and Engelmann, Francis},
  booktitle = {NeurIPS},
  year      = {2023},
  eprint    = {https://arxiv.org/abs/2306.13631v2}
}

@inproceedings{assran2023jpea,
  title={Self-Supervised Learning from Images with a Joint-Embedding Predictive Architecture},
  author={Assran, Mahmoud and Duval, Quentin and Misra, Ishan and Bojanowski, Piotr and Vincent, Pascal and Rabbat, Michael and LeCun, Yann and Ballas, Nicolas},
  booktitle={CVPR},
  year={2023}
}

@article{barsalou2008grounded,
  title={Grounded cognition},
  author={Barsalou, Lawrence W},
  journal={Annual Review of Psychology},
  year={2008}
}

@article{shams2008benefits,
  title={Benefits of multisensory learning},
  author={Shams, Ladan and Seitz, Aaron R},
  journal={Trends in cognitive sciences},
  year={2008}
}

@article{lecun2022path,
  title={A path towards autonomous machine intelligence version 0.9.2},
  author={LeCun, Yann},
  journal={OpenReview},
  year={2022}
}

@inproceedings{zhou2018re10k,
  title={Stereo Magnification: Learning View Synthesis using Multiplane Images},
  author={Zhou, Tinghui and Tucker, Richard and Flynn, John and Fyffe, Graham and Snavely, Noah},
  Booktitle={SIGGRAPH},
  year={2018}
}

@inproceedings{wang2025vggt,
  title={Vggt: Visual geometry grounded transformer},
  author={Wang, Jianyuan and Chen, Minghao and Karaev, Nikita and Vedaldi, Andrea and Rupprecht, Christian and Novotny, David},
  booktitle={CVPR},
  year={2025}
}

@inproceedings{radford2021clip,
  title={Learning transferable visual models from natural language supervision},
  author={Radford, Alec and Kim, Jong Wook and Hallacy, Chris and Ramesh, Aditya and Goh, Gabriel and Agarwal, Sandhini and Sastry, Girish and Askell, Amanda and Mishkin, Pamela and Clark, Jack and others},
  booktitle={ICML},
  year={2021},
}

@inproceedings{li2022lseg,
  title={Language-driven Semantic Segmentation},
  author={Boyi Li and Kilian Q Weinberger and Serge Belongie and Vladlen Koltun and Rene Ranftl},
  booktitle={ICLR},
  year={2022},
}

@inproceedings{hu2022lora,
  title={Lora: Low-rank adaptation of large language models.},
  author={Hu, Edward J and Shen, Yelong and Wallis, Phillip and Allen-Zhu, Zeyuan and Li, Yuanzhi and Wang, Shean and Wang, Lu and Chen, Weizhu and others},
  booktitle={ICLR},
  year={2022}
}

@inproceedings{kobayashi2022decomposing,
  title={Decomposing nerf for editing via feature field distillation},
  author={Kobayashi, Sosuke and Matsumoto, Eiichi and Sitzmann, Vincent},
  booktitle={NeurIPS},
  year={2022}
}

@inproceedings{chen2023clip2scene,
  title={Clip2scene: Towards label-efficient 3d scene understanding by clip},
  author={Chen, Runnan and Liu, Youquan and Kong, Lingdong and Zhu, Xinge and Ma, Yuexin and Li, Yikang and Hou, Yuenan and Qiao, Yu and Wang, Wenping},
  booktitle={CVPR},
  year={2023}
}

@article{arnaud2025locate3d,
  title={Locate 3D: Real-World Object Localization via Self-Supervised Learning in 3D},
  author={Arnaud, Sergio and McVay, Paul and Martin, Ada and Majumdar, Arjun and Jatavallabhula, Krishna Murthy and Thomas, Phillip and Partsey, Ruslan and Dugas, Daniel and Gejji, Abha and Sax, Alexander and others},
  journal={arXiv:2504.14151},
  year={2025}
}

@article{zeid2025dinoinroom,
  title={DINO in the Room: Leveraging 2D Foundation Models for 3D Segmentation},
  author={Zeid, Karim Abou and Yilmaz, Kadir and de Geus, Daan and Hermans, Alexander and Adrian, David and Linder, Timm and Leibe, Bastian},
  journal={arXiv:2503.18944},
  year={2025}
}

@article{jain2025univlg,
  title={Unifying 2D and 3D Vision-Language Understanding},
  author={Jain, Ayush and Swerdlow, Alexander and Wang, Yuzhou and Arnaud, Sergio and Martin, Ada and Sax, Alexander and Meier, Franziska and Fragkiadaki, Katerina},
  journal={arXiv:2503.10745},
  year={2025}
}

@article{fan2024lsm,
  title={Large spatial model: End-to-end unposed images to semantic 3d},
  author={Fan, Zhiwen and Zhang, Jian and Cong, Wenyan and Wang, Peihao and Li, Renjie and Wen, Kairun and Zhou, Shijie and Kadambi, Achuta and Wang, Zhangyang and Xu, Danfei and others},
  journal={Advances in neural information processing systems},
  volume={37},
  pages={40212--40229},
  year={2024}
}

@article{tschannen2025siglip2,
  title={Siglip 2: Multilingual vision-language encoders with improved semantic understanding, localization, and dense features},
  author={Tschannen, Michael and Gritsenko, Alexey and Wang, Xiao and Naeem, Muhammad Ferjad and Alabdulmohsin, Ibrahim and Parthasarathy, Nikhil and Evans, Talfan and Beyer, Lucas and Xia, Ye and Mustafa, Basil and others},
  journal={arXiv preprint arXiv:2502.14786},
  year={2025}
}

@inproceedings{ranzinger2024radio,
  title={Am-radio: Agglomerative vision foundation model reduce all domains into one},
  author={Ranzinger, Mike and Heinrich, Greg and Kautz, Jan and Molchanov, Pavlo},
  booktitle={Proceedings of the IEEE/CVF Conference on Computer Vision and Pattern Recognition},
  pages={12490--12500},
  year={2024}
}

@article{puig2023habitat,
  title={Habitat 3.0: A co-habitat for humans, avatars and robots},
  author={Puig, Xavier and Undersander, Eric and Szot, Andrew and Cote, Mikael Dallaire and Yang, Tsung-Yen and Partsey, Ruslan and Desai, Ruta and Clegg, Alexander William and Hlavac, Michal and Min, So Yeon and others},
  journal={arXiv preprint arXiv:2310.13724},
  year={2023}
}

@article{conceptfusion,
  author    = {Jatavallabhula, {Krishna Murthy} and Kuwajerwala, Alihusein and Gu, Qiao and Omama, Mohd and Chen, Tao and Li, Shuang and Iyer, Ganesh and Saryazdi, Soroush and Keetha, Nikhil and Tewari, Ayush and Tenenbaum, {Joshua B.} and {de Melo}, {Celso Miguel} and Krishna, Madhava and Paull, Liam and Shkurti, Florian and Torralba, Antonio},
  title     = {ConceptFusion: Open-set Multimodal 3D Mapping},
  journal   = {Robotics: Science and Systems (RSS)},
  year      = {2023},
}

@article{huang2024openins3d,
      title={OpenIns3D: Snap and Lookup for 3D Open-vocabulary Instance Segmentation}, 
      author={Zhening Huang and Xiaoyang Wu and Xi Chen and Hengshuang Zhao and Lei Zhu and Joan Lasenby},
      journal={European Conference on Computer Vision},
      year={2024}
    }

@article{Zhang_2025,
   title={Self-Supervised Learning of LiDAR 3D Point Clouds Via 2D-3D Neural Calibration},
   ISSN={1939-3539},
   url={http://dx.doi.org/10.1109/TPAMI.2025.3584625},
   DOI={10.1109/tpami.2025.3584625},
   journal={IEEE Transactions on Pattern Analysis and Machine Intelligence},
   publisher={Institute of Electrical and Electronics Engineers (IEEE)},
   author={Zhang, Yifan and Hou, Junhui and Ren, Siyu and Wu, Jinjian and Yuan, Yixuan and Shi, Guangming},
   year={2025},
   pages={1–17} }
}

\newpage
\appendix
\section*{Appendix}
\begin{figure}
    \centering
    \includegraphics[width=1.0\textwidth]{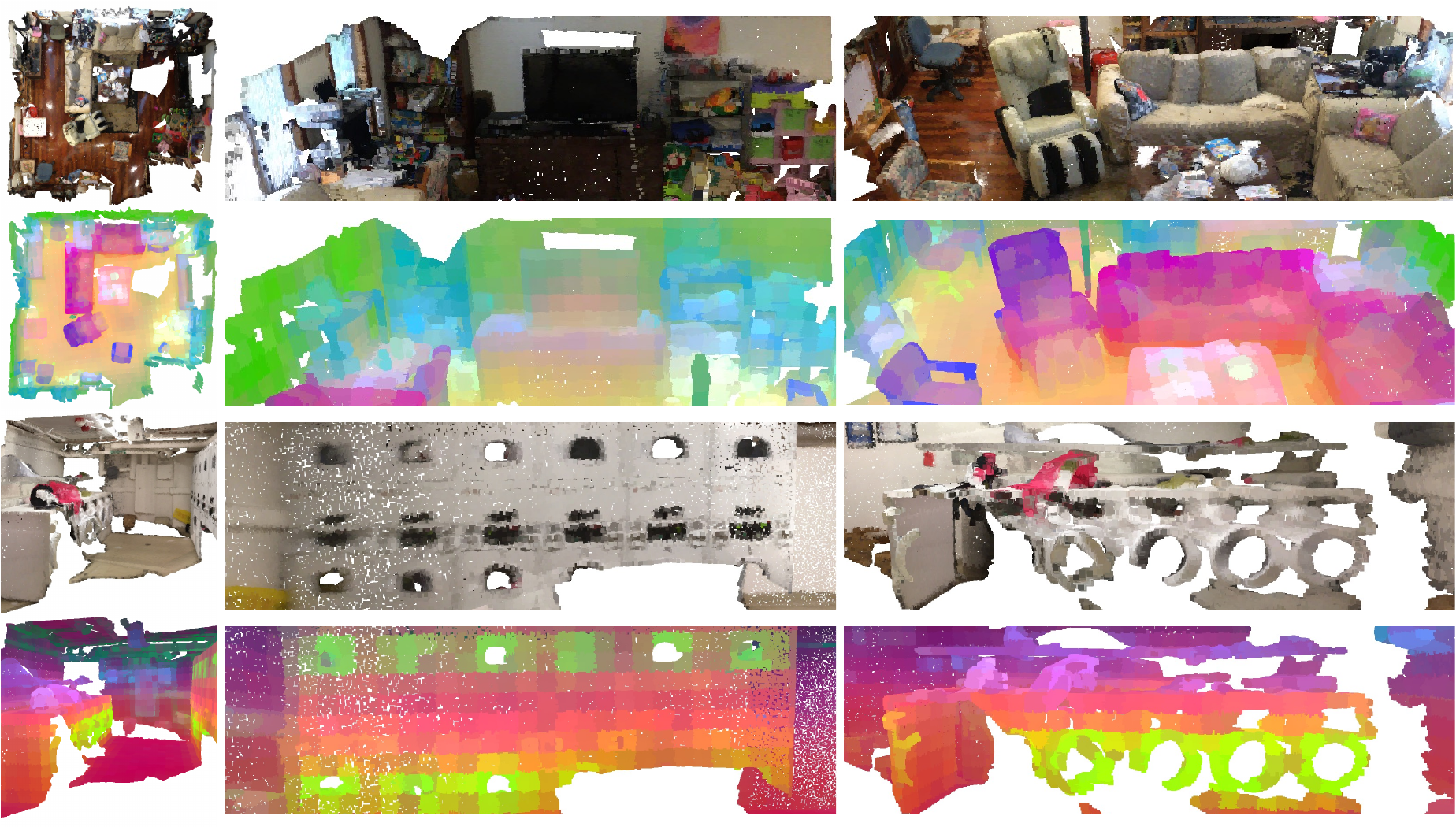}
    \caption{\textbf{Qualitative visualization.} Concerto performs well across different point cloud inputs: a complete scene (top two rows) and an incomplete scene (bottom two rows).}
    \vspace{-2mm}
    \label{fig:scene1appendix}
\end{figure}

Concerto is a superior spatial representation point encoder capable of handling a wide range of scene types, including those with varying completeness in \figref{fig:scene1appendix}, video-lifted point clouds in \figref{fig:videoappendix}, and the large scene in \figref{fig:lanlocappendix}. Here, we further present the detailed implementation and results.

\section{Additional Implementation}
We adopt the detailed parameters from Sonata~\cite{wu2025sonata} for intra-modal self-distillation and refer readers to the original Sonata paper for an in-depth description of its implementation. In this section, we provide a thorough explanation of the implementation for cross-modal joint embedding prediction.

\subsection{Combination of Intra-Modal and Cross-Modal Learning}
As Sonata, we use 4 local views, 2 masked views, and 2 global views, with the first global view serving as the principal view. For cross-modal joint embedding prediction, we utilize the representations from the first masked view (based on the principal view) to predict the corresponding image representations. The cross-modal cosine similarity loss is computed at upcast level 3, while the online clustering cross-entropy loss for intra-modal self-distillation is calculated at upcast level 2.

\subsection{Correspondence Between Pixels and Points}
To establish reliable 3D point to 2D pixel correspondences across camera views, we employ a two-step approach: 3D-to-2D projection followed by depth-based visibility verification.

Let $\vp = (X, Y, Z)^T$ denote a 3D point in world coordinates. Each camera $c$ is defined by intrinsic matrix $\mK$ and extrinsic matrix $[\mR|\vt]$. The standard pinhole camera model projects the 3D point $\vp$ to 2D pixel coordinates $(x,y)$ and a projected depth $d_{\mathrm{proj}}$:
\begin{equation}
d_{\mathrm{proj}} \begin{bmatrix} x \\ y \\ 1 \end{bmatrix} = \mK [\mR|\vt] \begin{bmatrix} X \\ Y \\ Z \\ 1 \end{bmatrix}. \label{eq:projection}
\end{equation}
To account for occlusions, we perform a visibility check comparing $d_{\mathrm{proj}}$ with the depth value $d_c = \mD_c(x,y)$ retrieved from camera $c$'s depth map $\mD_c$ at the projected pixel coordinate $(x, y)$. The point is considered visible if:
\begin{equation}
|d_c - d_{\mathrm{proj}}| < \epsilon_{\mathrm{depth}}, \label{eq:depth_check}
\end{equation}
where $\epsilon_{\mathrm{depth}}$ is set to 0.01 in our experiments. Additionally, the correspondence is rejected if $(x,y)$ falls outside image bounds or $\mD_c(x,y)$ contains invalid depth. This visibility check establishes a mapping between 3D points and corresponding 2D pixels, enabling direct correspondence between 3D points and ViT patches for cross-model joint embedding prediction mechanisms. Depending on the dataset, the depth map $\mD_c$ is obtained in different ways:

\begin{itemize}
    \item \textbf{RGBD datasets.} Depth maps are directly available as the depth channel of RGBD images, such as Structured3D~\cite{zheng2020structured3d}.
    \item \textbf{Known ground truth mesh.} For datasets like ScanNet~\cite{dai2017scannet}, ScanNet++~\cite{yeshwanth2023scannet++}, S3DIS~\cite{armeni2016s3dis}, and ARKitScenes~\cite{dehghan2021arkitscenes}, depth maps are rendered from the ground truth 3D mesh using camera parameters.
    \item \textbf{Pixel-aligned point clouds.} For video-lifted point clouds (e.g., using VGGT~\cite{wang2025vggt} on RealEstate10K~\cite{zhou2018re10k}), per-view depth maps $\mD_i$ are generated alongside point clouds $\sP_i$. A point $\vp \in \sP_i$ from camera $i$ can be visible from camera $j$ if it passes the visibility check. 
\end{itemize}

For HM3D~\cite{ramakrishnan2021hm3d}, which does not provide the raw images, we leverage Habitat-Sim~\cite{puig2023habitat} to simulate the scenes. For each navigatable room, we capture four images around the room with random initial camera orientations. The angular difference between consecutive images is 90 degrees. We record the camera parameters to compute the correspondence between points and pixels, as described previously.
The total collections of our training data are shown in \tabref{tab:dataimg} and \tabref{tab:datapcd}.

\begin{table*}[!t]
    \begin{minipage}{1.0\textwidth}
    \centering
        \tablestyle{1pt}{1.08}
        \begin{tabular}{p{25mm}|>{\raggedleft\arraybackslash}p{15mm}|>{\raggedleft\arraybackslash}p{15mm}>{\raggedleft\arraybackslash}p{15mm}>{\raggedleft\arraybackslash}p{15mm}>{\raggedleft\arraybackslash}p{15mm}}\toprule
Dataset &Source &Train &Val &Test &All \\\midrule
ScanNet~\cite{dai2017scannet} &real &26,428 &7,354 &2,877 &36,659 \\
ScanNet++~\cite{yeshwanth2023scannet++} &real &49,315 &1,583 &1,208 &52,106 \\
S3DIS~\cite{armeni2016s3dis} &real &10,977 &3,668 &0 &14,645 \\
ArkitScenes~\cite{dehghan2021arkitscenes} &real &72,481 &9,786 &0 &82,267 \\
HM3D~\cite{ramakrishnan2021hm3d} &real &64,936 &8,240 &0 &73,176 \\
Structured3D~\cite{zheng2020structured3d} &synthesis &65,160 &6,722 &6,396 &78,278 \\
RE10K~\cite{zhou2018re10k} &real &166,680 &0 &18,464 &185,144 \\\cmidrule(lr){1-6}
\cellcolor[HTML]{efefef}Concerto~(ours) &\cellcolor[HTML]{efefef}mixed &\cellcolor[HTML]{efefef}\textbf{455,972} &\cellcolor[HTML]{efefef}\textbf{37,353} &\cellcolor[HTML]{efefef}\textbf{28,945} &\cellcolor[HTML]{efefef}\textbf{522,270} \\
\bottomrule
\end{tabular}

        \vspace{-2mm}
        \caption{\textbf{Image Data Source Collection.}}
        \label{tab:dataimg}
        \vspace{1mm}
    \vspace{3mm}
        \tablestyle{1pt}{1.08}
        \begin{tabular}{p{25mm}|>{\raggedleft\arraybackslash}p{15mm}|>{\raggedleft\arraybackslash}p{15mm}>{\raggedleft\arraybackslash}p{15mm}>{\raggedleft\arraybackslash}p{15mm}>{\raggedleft\arraybackslash}p{15mm}}\toprule
Dataset &Source &Train &Val &Test &All \\\midrule
ScanNet~\cite{dai2017scannet} &real &1,201 &312 &100 &1,613 \\
ScanNet++~\cite{yeshwanth2023scannet++} &real &856 &50 &50 &956 \\
S3DIS~\cite{armeni2016s3dis} &real &204 &68 &0 &272 \\
ArkitScenes~\cite{dehghan2021arkitscenes} &real &4,498 &549 &0 &5,047 \\
HM3D~\cite{ramakrishnan2021hm3d} &real &8,117 &1,030 &0 &9,147 \\
Structured3D~\cite{zheng2020structured3d} &synthesis &16,635 &1,722 &1,648 &20,005 \\
RE10K~\cite{zhou2018re10k} &real &41,670 &0 &4,612 &46,282 \\\cmidrule(lr){1-6}
\cellcolor[HTML]{efefef}Concerto~(ours) &\cellcolor[HTML]{efefef}mixed &\cellcolor[HTML]{efefef}\textbf{74,894} &\cellcolor[HTML]{efefef}\textbf{3,785} &\cellcolor[HTML]{efefef}\textbf{6,459} &\cellcolor[HTML]{efefef}\textbf{85,138} \\
\bottomrule
\end{tabular}

        \vspace{-2mm}
        \caption{\textbf{Point Cloud Data Source Collection.}}
        \label{tab:datapcd}
        \vspace{1mm}
    \end{minipage}
\end{table*}

\subsection{Image Augmentations}
We implement the same point cloud augmentations as Sonata. For image augmentations, we initially adopt the process from DINOv2~\cite{oquab2023dinov2}, excluding geometric augmentations to simplify the alignment between pixels and points. Specifically, we apply color jittering, random grayscale, and Gaussian blur to the images, consistent with the settings used in DINOv2. This results in a slight drop in the mIoU on ScanNet semantic segmentation to 75.27\%, compared to using the original images. Consequently, we continue to explore more suitable image augmentations. In the ablation study, we apply random color jittering, with the same intensity as the point cloud augmentations, along with Gaussian blur. This weaker augmentation improves Concerto's performance, which is expected since the image encoder is currently frozen. Stronger augmentations may yield better results once both the image and point branches are unlocked for joint learning.

\vspace{-3mm}
\subsection{Experimental Setting}
\mypara{Software and hardware environment.}
\begin{itemize}
    \item CUDA version: 12.4
    \item PyTorch version: 2.4.1
    \item Python version: 3.10.15
    \item GPU: Nvidia H20 $\times$ 16 for pretraining; Nvidia H20 $\times$ 8 for evaluation.
    \item CPU: $\times$ 360 for pretraining; $\times$ 180 for evaluation.
    \item Memory: 3600GB for pretraining; 1800GB for evaluation.
    \item Time: 85h for pretraining of base model without video data.
\end{itemize}

\mypara{Data license.}
We use the open-source datasets ScanNet~\cite{dai2017scannet}, ScanNet++~\cite{yeshwanth2023scannet++}, S3DIS~\cite{armeni2016s3dis}, Structured3D~\cite{zheng2020structured3d}, ARKitScenes~\cite{dehghan2021arkitscenes}, Habitat Matterport3D~\cite{ramakrishnan2021hm3d} and RealEstate10K~\cite{zhou2018re10k} in latest versions. S3DIS, Structured3D, ScanNet, and ScanNet++ have custom licenses. RealEstate10K is licensed by Google LLC under a Creative Commons Attribution 4.0 International License. ARKitScenes is licensed by Apple Inc. HM3D is licensed by Matterport.

\mypara{Training details.}
For pretraining, we leverage all train, val, and test splits to train the self-supervised model. For evaluation with linear probing, decoder probing, and full fine-tuning, we train on the train split and test on the val split of ScanNet, ScanNet++, ScanNet200, and Area 5 of S3DIS. We use AdamW as the optimizer, and cosine annealing policy as the scheduler. The learning rate is adjusted with the encoder depth, and the max one is 0.004. The pretraining epoch is 100. For cross-modal joint embedding prediction, we set DINOv2 image encoder input resolution 518$\times$518.

\section{Additional Results}
\begin{figure}
    \centering
    \includegraphics[width=1.0\textwidth]{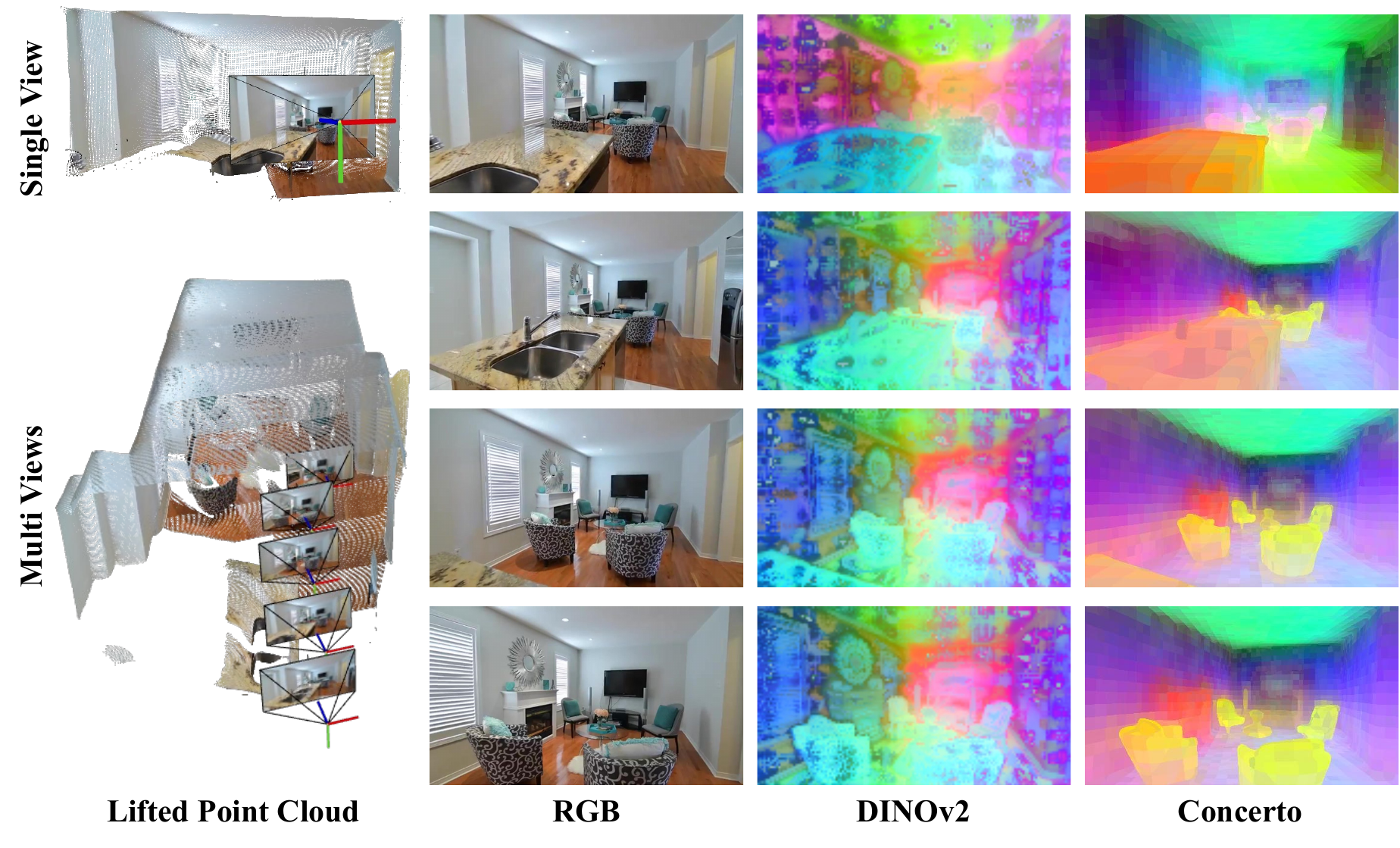}
    \caption{\textbf{Video perception.} Concerto can be applied to single-view (top row) and multi-view video-lifted data (bottom three rows). We visualize the PCA of one video in RE10K~\cite{zhou2018re10k}. In the multi-view setting, the representations from all the frames are computed together for consistency.}
    \vspace{-3mm}
    \label{fig:videoappendix}
\end{figure}
\subsection{Concerto with Video-Lifted Point Clouds}

We utilize the current feed-forward reconstruction model VGGT~\cite{wang2025vggt} to lift RealEstate10K~\cite{zhou2018re10k} video data to point clouds. Based on the camera poses, we heuristically select video clips with larger camera pose transforms in comparison and abandon those with smaller camera pose transforms. With these video clips, we can build a video dataset with more completed scenes. In \figref{fig:videoappendix}, we utilize Concerto to deal with single-view lifted data and multi-view lifted data. The visualizations show that Concerto adapts well to these two situations, suggesting that Concerto cannot only be applied to the offline video reconstruction but also to the single-view forward situation.

\begin{figure}
    \centering
    \includegraphics[width=1.0\textwidth]{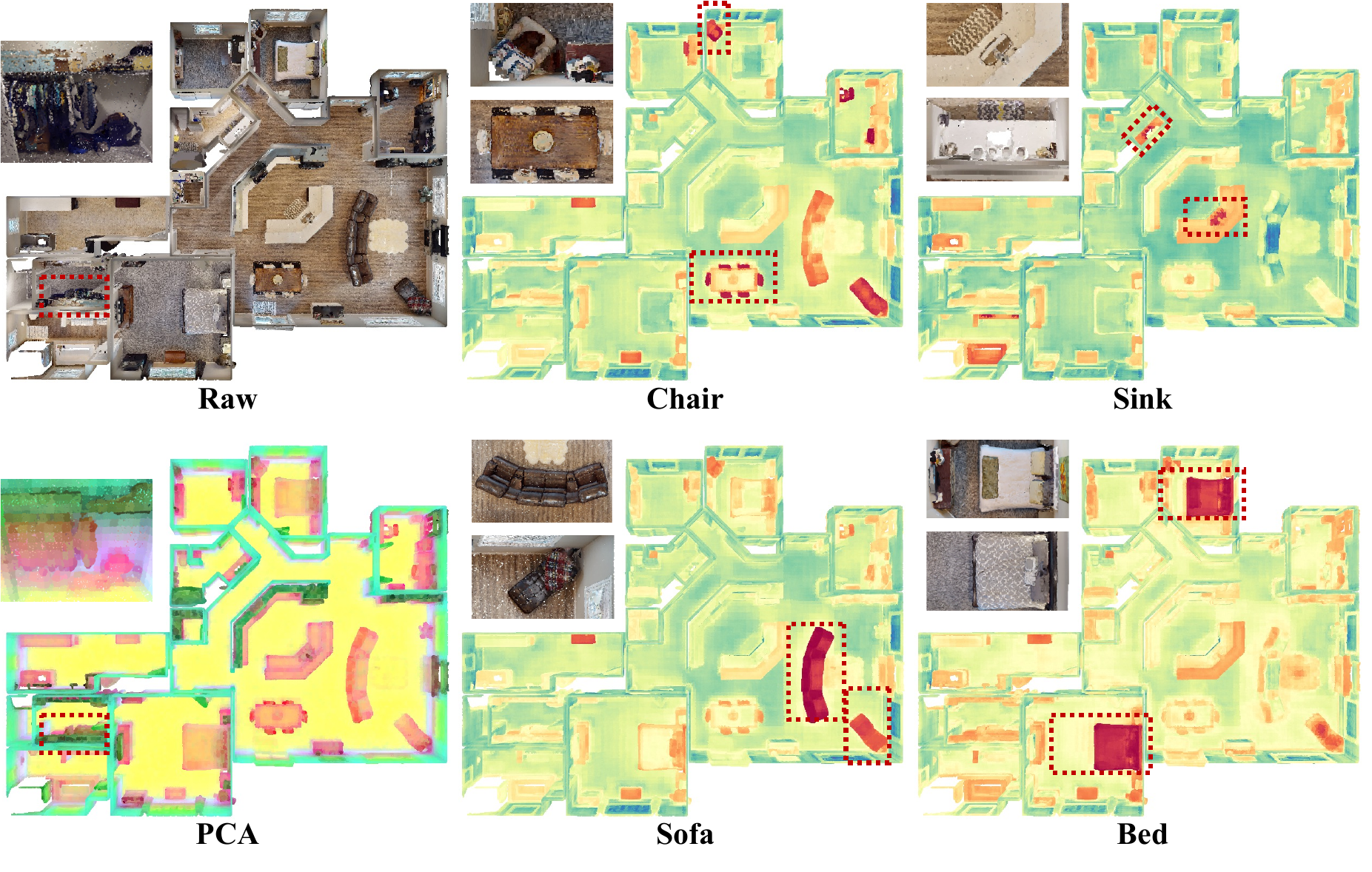}
    \vspace{-6mm}
    \caption{\textbf{Language locate.} We visualize the PCA of a large house scene from HM3D~\cite{ramakrishnan2021hm3d} along with the heatmap of zero-shot language-based object localization results. The upper-left part of the scene shows detailed local information. Given specific words, Concerto with text-aligned linear probing successfully locates objects in a zero-shot setting.}
    \label{fig:lanlocappendix}
\end{figure}

\subsection{Concerto with Language Probing}
We leverage a simple linear layer to translate the representations from Concerto to CLIP's text space. During training, we force the linear probing output to align with the LSeg~\cite{li2022lseg} image encoder's output, which does not need the ground truth labels to supervise. In the aligning process, we do not use masks and crop augmentations. The visualization results are shown in \figref{fig:lanlocappendix}.

\begin{table*}[!t]
    \centering
    \tablestyle{4.4pt}{1.08}
    \begin{tabular}{lrrrrrrrrrrrrr}\toprule
Model &\multicolumn{3}{c}{ScanNet Val} &\multicolumn{3}{c}{ScanNet200 Val} &\multicolumn{3}{c}{ScanNet++ Val} &\multicolumn{3}{c}{S3DIS Area 5} \\\cmidrule(lr){1-1}\cmidrule(lr){2-4}\cmidrule(lr){5-7}\cmidrule(lr){8-10}\cmidrule(lr){11-13}
img. enc. &mIoU &mAcc &allAcc &mIoU &mAcc &allAcc &mIoU &mAcc &allAcc &mIoU &mAcc &allAcc \\\midrule
\cellcolor[HTML]{f3f7fc}DINOv2 (lin.) &\cellcolor[HTML]{def3e2}\textbf{77.3} &\cellcolor[HTML]{def3e2}\textbf{86.6} &\cellcolor[HTML]{def3e2}\textbf{91.7} &\cellcolor[HTML]{def3e2}\textbf{37.4} &\cellcolor[HTML]{def3e2}\textbf{49.5} &\cellcolor[HTML]{def3e2}\textbf{83.3} &\cellcolor[HTML]{eef9ee}45.7 &\cellcolor[HTML]{eef9ec}60.5 &\cellcolor[HTML]{eef9ec}86.5 &\cellcolor[HTML]{def3e2}\textbf{73.5} &\cellcolor[HTML]{def3e2}\textbf{81.3} &90.9 \\
SigLIP2 (lin.) &\cellcolor[HTML]{eef9ee}76.3 &\cellcolor[HTML]{eef9ee}86.0 &\cellcolor[HTML]{eef9ee}91.4 &\cellcolor[HTML]{eef9ee}36.7 &\cellcolor[HTML]{eef9ee}48.9 &\cellcolor[HTML]{eef9ee}82.7 &\cellcolor[HTML]{def3e0}\textbf{45.8} &\cellcolor[HTML]{def3e0}\textbf{61.4} &\cellcolor[HTML]{def3e0}\textbf{86.8} &72.3 &79.4 &\cellcolor[HTML]{def3e2}\textbf{91.0} \\
RADIO (lin.) &73.5 &84.0 &90.3 &31.0 &42.3 &81.8 &42.7 &57.2 &85.3 &\cellcolor[HTML]{eef9ee}72.9 &\cellcolor[HTML]{eef9ee}80.5 &\cellcolor[HTML]{eef9ee}90.9 \\\cmidrule(lr){1-13}
\cellcolor[HTML]{f3f7fc}DINOv2 (dec.) &\cellcolor[HTML]{def3e0}\textbf{79.5} &\cellcolor[HTML]{def3e0}\textbf{87.6} &\cellcolor[HTML]{def3e0}\textbf{92.6} &\cellcolor[HTML]{def3e0}\textbf{37.8} &\cellcolor[HTML]{def3e0}\textbf{50.5} &\cellcolor[HTML]{def3e0}\textbf{84.1} &\cellcolor[HTML]{def3e0}\textbf{48.3} &\cellcolor[HTML]{def3e0}\textbf{62.3} &\cellcolor[HTML]{def3e0}\textbf{87.7} &\cellcolor[HTML]{def3e0}\textbf{75.5} &\cellcolor[HTML]{def3e0}\textbf{84.2} &\cellcolor[HTML]{def3e0}\textbf{92.3} \\
SigLIP2 (dec.) &\cellcolor[HTML]{eef9ee}78.8 &\cellcolor[HTML]{eef9ee}87.0 &\cellcolor[HTML]{eef9ee}92.4 &\cellcolor[HTML]{eef9ee}37.5 &\cellcolor[HTML]{eef9ee}47.7 &\cellcolor[HTML]{eef9ee}83.7 &\cellcolor[HTML]{eef9ee}46.8 &\cellcolor[HTML]{eef9ee}58.1 &\cellcolor[HTML]{eef9ee}87.0 &73.6 &79.8 &91.3 \\
RADIO (dec.) &77.9 &85.7 &92.3 &33.9 &44.6 &83.4 &44.9 &56.5 &86.2 &\cellcolor[HTML]{eef9ec}74.8 &\cellcolor[HTML]{eef9ec}81.2 &\cellcolor[HTML]{eef9ec}92.2 \\\cmidrule(lr){1-13}
\cellcolor[HTML]{f3f7fc}DINOv2 (full.) &\cellcolor[HTML]{def3de}\textbf{80.7} &\cellcolor[HTML]{def3de}\textbf{87.4} &\cellcolor[HTML]{def3de}\textbf{93.1} &\cellcolor[HTML]{def3de}\textbf{39.2} &\cellcolor[HTML]{def3de}\textbf{50.2} &\cellcolor[HTML]{def3de}\textbf{85.0} &\cellcolor[HTML]{def3de}\textbf{50.7} &\cellcolor[HTML]{def3de}\textbf{63.3} &\cellcolor[HTML]{eef9ec}87.9 &\cellcolor[HTML]{def3de}\textbf{77.4} &\cellcolor[HTML]{def3de}\textbf{85.0} &\cellcolor[HTML]{def3de}\textbf{93.2} \\
SigLIP2 (full.) &\cellcolor[HTML]{eef9ee}79.7 &\cellcolor[HTML]{eef9ee}86.9 &92.7 &\cellcolor[HTML]{eef9ee}38.4 &\cellcolor[HTML]{eef9ee}49.9 &\cellcolor[HTML]{eef9ee}83.9 &\cellcolor[HTML]{eef9ee}50.0 &\cellcolor[HTML]{eef9ee}62.0 &88.2 &75.0 &80.2 &92.5 \\
RADIO (full.) &79.6 &86.6 &\cellcolor[HTML]{eef9ee}92.7 &36.1 &46.9 &83.8 &48.4 &60.6 &\cellcolor[HTML]{def3dc}\textbf{88.2} &\cellcolor[HTML]{eef9ec}75.1 &\cellcolor[HTML]{eef9ec}80.5 &\cellcolor[HTML]{eef9ec}92.8 \\
\bottomrule
\end{tabular}

    \vspace{-2mm}
    \caption{\textbf{Segmentic segmentation of Concerto with different image encoders.} Concerto with DINOv2 based on self-distillation has the best performance in general. }
    \label{tab:imgenc}
    \vspace{-4mm}
\end{table*}
\subsection{Results with Different 2D Encoder}
In this section, we compare the performance of different strong image encoders: DINOv2~\cite{oquab2023dinov2}, SigLIPv2~\cite{tschannen2025siglip2}, and RADIO~\cite{ranzinger2024radio}. We adopt DINOv2 L version with a resolution of 518$\times$518, SigLIPv2 So400m version with a patch size of 16 and resolution 512$\times$512, and RADIOv2.5 L version with a resolution of 768$\times$768. For each model, we pretrain a variant of Concerto on 40k data, excluding video-lifted data. We evaluate these models across four datasets on semantic segmentation, as shown in \tabref{tab:imgenc}. The results reveal that the Concerto model based on DINOv2, using self-distillation, achieves the highest mIoU in general. This suggests that in our joint self-supervised learning framework, the optimal synergy is achieved when representations from different domains are derived through intra-modal self-distillation. RADIO, which incorporates distilled information from multiple models, may damage the original self-distillation features from DINOv2, thus leading to a decrease in performance. 

\vspace{-1mm}
\subsection{Results with LoRA Finetuning}

\begin{table*}[!t]
    \centering
    \tablestyle{5.3pt}{1.08}
    \begin{tabular}{y{22.5mm}rz{8mm}cccccccccccc}\toprule
Data Efficiency &\multicolumn{2}{c}{Params} &\multicolumn{5}{c}{Limited Scenes (Pct.)} &\multicolumn{5}{c}{Limited Annotation (Pts.)} \\\cmidrule(lr){1-1} \cmidrule(lr){2-3} \cmidrule(lr){4-8} \cmidrule(lr){9-13}
Methods &Learn. &Pct. &1\% &5\% &10\% &20\% &100\% &20 &50 &100 &200 &Full \\\midrule
Concerto (lin.) &0.02M &0.02\% &\cellcolor[HTML]{eef9f0}48.2 &\cellcolor[HTML]{eef9f0}69.1 &73.6 &75.0 &\textcolor{darkgray}{77.3} &\cellcolor[HTML]{eef9ee}73.9 &75.2 &76.2 &76.3 &\textcolor{darkgray}{77.3} \\
Concerto (dec.) &16.3M &13.1\% &44.6 &67.9 &73.7 &74.6 &\textcolor{darkgray}{79.5} &72.6 &74.6 &76.7 &77.6 &\textcolor{darkgray}{79.5} \\
Concerto (full.) &124.8M &100.0\% &46.5 &69.0 &\cellcolor[HTML]{def3e0}\textbf{75.3} &\cellcolor[HTML]{eef9ee}76.1 &\textcolor{darkgray}{80.7} &73.3 &\cellcolor[HTML]{eef9f0}76.7 &\cellcolor[HTML]{eef9f0}77.6 &\cellcolor[HTML]{eef9f0}78.4 &\textcolor{darkgray}{80.7} \\
Concerto (lora) &0.3M &0.2\% &\cellcolor[HTML]{def3e2}\textbf{48.4} &\cellcolor[HTML]{def3e2}\textbf{70.2} &\cellcolor[HTML]{eef9f0}74.9 &\cellcolor[HTML]{def3e2}\textbf{76.8} &\textcolor{darkgray}{79.8} &\cellcolor[HTML]{def3e2}\textbf{75.1} &\cellcolor[HTML]{def3e2}\textbf{77.2} &\cellcolor[HTML]{def3e2}\textbf{78.3} &\cellcolor[HTML]{def3e2}\textbf{78.7} &\textcolor{darkgray}{79.8} \\
\bottomrule
\end{tabular}

    \vspace{-2mm}
    \caption{\textbf{Parameter Efficiency with LoRA.} Concerto with LoRA significantly improves the performance with a minimal number of learnable parameters, highlighting the reliability of pretrained Concerto representations and the effectiveness of LoRA fine-tuning.}
    \label{tab:loraefficiency}
    \vspace{-2mm}
\end{table*}
\begin{table*}[!t]
    \centering
    \tablestyle{2.1pt}{1.08}
    \begin{tabular}{lrrrrrrrrrrrrrrr}\toprule
LoRA &\multicolumn{2}{c}{Params} &\multicolumn{3}{c}{ScanNet Val} &\multicolumn{3}{c}{ScanNet200 Val} &\multicolumn{3}{c}{ScanNet++ Val} &\multicolumn{3}{c}{S3DIS Area 5} \\\cmidrule(lr){1-1} \cmidrule(lr){2-3} \cmidrule(lr){4-6} \cmidrule(lr){7-9} \cmidrule(lr){10-12} \cmidrule(lr){13-15}
Methods &Learn. &Pct. &mIoU &mAcc &allAcc &mIoU &mAcc &allAcc &mIoU &mAcc &allAcc &mIoU &mAcc &allAcc \\\midrule
Concerto (lin.) &\tinyless0.2M &\tinyless0.2\% &77.3 &86.6 &91.7 &37.4 &49.5 &83.3 &45.6 &60.5 &86.5 &73.5 &81.3 &90.9 \\
Concerto (dec.) &16.3M &13.1\% &79.5 &\cellcolor[HTML]{eef9ee}87.6 &92.6 &37.8 &\cellcolor[HTML]{eef9ec}50.5 &84.1 &\cellcolor[HTML]{eef9ee}48.3 &\cellcolor[HTML]{eef9ee}62.3 &\cellcolor[HTML]{eef9ee}87.7 &75.5 &\cellcolor[HTML]{eef9ec}84.2 &92.3 \\
Concerto (full.) &124.8M &100.0\% &\cellcolor[HTML]{def3e2}\textbf{80.7} &87.4 &\cellcolor[HTML]{def3e2}\textbf{93.1} &\cellcolor[HTML]{def3e2}\textbf{39.2} &50.2 &\cellcolor[HTML]{def3e2}\textbf{85.0} &\cellcolor[HTML]{def3e0}\textbf{50.7} &\cellcolor[HTML]{def3e0}\textbf{63.3} &\cellcolor[HTML]{def3e0}\textbf{87.9} &\cellcolor[HTML]{def3de}\textbf{77.4} &\cellcolor[HTML]{def3e0}\textbf{85.0} &\cellcolor[HTML]{def3e0}\textbf{93.2} \\
Concerto (lora) &\tinyless0.5M &\tinyless0.5\% &\cellcolor[HTML]{eef9ee}79.8 &\cellcolor[HTML]{def3e2}\textbf{87.9} &\cellcolor[HTML]{eef9ee}92.7 &\cellcolor[HTML]{eef9ee}38.4 &\cellcolor[HTML]{def3e2}\textbf{51.9} &\cellcolor[HTML]{eef9ee}84.1 &47.3 &60.8 &87.7 &\cellcolor[HTML]{eef9ec}75.5 &81.4 &\cellcolor[HTML]{eef9ec}92.6 \\
\bottomrule
\end{tabular}

    \vspace{-2mm}
    \caption{\textbf{Semantic segmentation with LoRA.} We compare the LoRA fine-tuning method on Concerto across four semantic segmentation benchmarks, demonstrating LoRA's remarkable capacity in general and the reliability of Concerto's original pretrained representations.}
    \label{tab:loracompare}
    \vspace{-4mm}
\end{table*}
From the main results, we observe that linear probing outperforms full-finetuning in extreme data-scarce scenarios. This suggests that training methods may benefit from shifting toward LoRA-based fine-tuning. In this section, we present the results of LoRA fine-tuning. Specifically, we adapt LoRA to the point encoder and evaluate it with linear probing. We set the LoRA rank to 8, the LoRA alpha to 16, and the dropout rate to 0.1. 

The results of ScanNet Data Efficiency are shown in \tabref{tab:loraefficiency}. These results demonstrate that the LoRA-based method outperforms both linear probing and full fine-tuning in terms of mIoU across most scenarios, despite a small increase in learnable parameters compared to the original linear probing. This suggests that LoRA is an effective fine-tuning approach, particularly when data is limited. Notably, linear probing with LoRA achieves performance comparable to decoder probing in the full evaluation and only a 0.9\% performance drop compared to full fine-tuning on mIoU, while offering significant improvements in training efficiency.

We also evaluate the LoRA fine-tuning on Concerto across four benchmarks, as shown in \tabref{tab:loracompare}. The results demonstrate that LoRA fine-tuning shows performance comparable to decoder probing, even with relatively small learnable parameters. Overall, the LoRA fine-tuning demonstrates strong efficiency and performance across various benchmarks, highlighting two key insights: Concerto already yields reliable and generalizable representations, and leveraging pretrained representations combined with LoRA fine-tuning is both efficient and effective for further task adaptation.

\end{document}